%% file: main.tex
\documentclass[11pt]{article}

\usepackage[final]{acl}

\usepackage{times}
\usepackage{latexsym}

\usepackage[T1]{fontenc}

\usepackage[utf8]{inputenc}
\usepackage{booktabs}
\usepackage{multirow}
\usepackage{amsmath}
\usepackage{hyperref}
\usepackage{subfig}
\usepackage{pgfplots}
\usepackage{float}
\usepackage{tikz}
\usepackage{float}
\usetikzlibrary{matrix,positioning,fit, calc}
\usepackage{xcolor}
\definecolor{pal1}{RGB}{220,235,250}
\definecolor{sentPos}{RGB}{180,230,180}
\definecolor{sentSPos}{RGB}{220,245,200}
\definecolor{sentNeu}{RGB}{240,240,200}

\usepackage{microtype}

\usepackage{inconsolata}

\usepackage{graphicx}

\usepackage{fvextra}
\usepackage{placeins}

\title{ 
Persona Prompting in Multimodal Urban Perception: Descriptive Convergence and Interpretive Variation}

\author{
  \textbf{Neemias B da Silva\textsuperscript{1,2}},
  \textbf{Matt Ratto\textsuperscript{1}},
  \textbf{Myriam Delgado\textsuperscript{2}},
  \textbf{Rodrigo Minetto\textsuperscript{2}},\\
  \textbf{Daniel Silver\textsuperscript{1}},
  \textbf{Thiago H Silva\textsuperscript{1,2}}\\
  \textsuperscript{1} University of Toronto, Toronto, Canada\\
  \textsuperscript{2} Universidade Tecnologica Federal do Parana, Curitiba, Brazil
\\
  \small{
    \textbf{Correspondence:} \href{neemias.buceli@mail.utoronto.ca}{neemias.buceli@mail.utoronto.ca}
  }
}

\begin{document}
\maketitle
\begin{abstract}
This study examines how persona prompting shapes language generated by two multimodal large language models in urban perception, a setting for examining subjective interpretations of shared visual evidence. We organize outputs into three functional layers: \emph{descriptive grounding} (captions), \emph{intermediate semantic layer} (perception tags), and \emph{interpretive framing} (justifications).  Using approximately 60,000 persona-conditioned annotations from each of two MLLMs, Qwen3-VL and Gemma4, we find that captions converge strongly across persona profiles and show only small attribute-associated differences. Justifications vary substantially more: economic status produces the largest difference in both models, with political orientation and personality also prominent. Paired image-level comparisons confirm larger justification than caption differences for these three attributes. For perception tags, personas sharing the same attribute level produce more similar tag sets than personas with different attribute levels, with the largest separation observed for economic status. Exploratory topic analysis further suggests persona-specific evaluative emphasis. Across models, profile-pair similarity patterns are strongly correlated for all three output types, although agreement is lowest for justifications. Overall, persona prompting affects interpretive framing more strongly than descriptive grounding.
\end{abstract}

\section{Introduction}\label{sec:intro}

Large language models (LLMs) are increasingly used as \emph{social simulators}, with persona prompts intended to approximate
diverse perspectives \cite{Apostolos2024,aherICML2023,Argyle_Busby_Fulda_Gubler_Rytting_Wingate_2023}. Applications include social-bias analysis and synthetic population simulation, raising questions about whether persona-conditioned models can represent collective perception and behavior \cite{park2024generativeagentsimulations1000,li2025simulating}. Although persona prompting may introduce pluralism, differences among persona-conditioned outputs need not reflect meaningful perspectives; they may instead stem from superficial linguistic variation or stereotypical model associations \cite{boelaert2025machine}.

Most prior work evaluates persona prompting through structured outputs, such as labels, choices, or task performance
\cite{hu-collier-2024-quantifying,beck-etal-2024-sensitivity,urbcom26-neemias}. Less is known about how personas shape generated
language, particularly descriptions and explanations. This distinction is especially relevant in multimodal settings, where visually grounded descriptions and subjective interpretations may respond differently to persona conditioning. Urban perception provides a useful setting: observers evaluate qualities such as safety, order, upkeep, beauty, and apparent socioeconomic status from visual cues, and these judgments vary across individuals and social contexts \cite{shen26,perceptSentPaper,dasilva2025multimodalllmssentiment}.

Despite growing evidence that persona conditioning can alter model behavior, an important question remains: \emph{where} do these effects emerge in multimodal generation? Differences between persona-conditioned outputs may reflect changes in what the model
describes, in the semantic concepts it associates with visual cues, or in how it interprets and evaluates the same evidence. Distinguishing these possibilities is important because differences in generated language may arise from description, semantic selection, or interpretation. We therefore examine persona effects across three complementary forms of output: descriptions of observable content, semantic perception tags associated with visual cues, and sentiment-linked justifications that interpret and evaluate the same visual evidence.

\begin{figure*}[t]
\centering

\begin{tikzpicture}[
    frameworkbox/.style={
        draw=black!55,
        rounded corners=4pt,
        text width=2.75cm,
        minimum height=1.18cm,
        inner sep=4pt,
        align=center,
        font=\footnotesize
    },
    inputbox/.style={
        draw=black!55,
        rounded corners=4pt,
        text width=2.0cm,
        minimum height=1.05cm,
        inner sep=4pt,
        align=center,
        fill=gray!10,
        font=\footnotesize
    },
    modelbox/.style={
        draw=black!55,
        rounded corners=4pt,
        text width=1.05cm,
        minimum height=1.05cm,
        inner sep=4pt,
        align=center,
        fill=gray!20,
        font=\footnotesize\bfseries
    },
    examplebox/.style={
        draw=black!50,
        rounded corners=5pt,
        text width=11.0cm,
        inner xsep=5pt,
        inner ysep=4pt,
        align=left,
        text=black!88,
        font=\scriptsize
    },
    flowline/.style={
        thick,
        gray!65,
        line cap=round,
        line join=round
    },
    flowarrow/.style={
        ->,
        thick,
        gray!65,
        line cap=round,
        line join=round,
        shorten >=1pt
    }
]


\begin{scope}[scale=0.90, transform shape]

\node[
    anchor=west,
    font=\small\bfseries
] at (-8.0,1.48)
{(a) Functional framework};

\node[inputbox] (input) at (-6.55,0.20) {%
    \textbf{Urban image}\\[1pt]
    \(+\)\\[-1pt]
    \textbf{Persona profile}
};

\node[
    modelbox,
    right=0.25cm of input
] (mllm) {MLLM};

\node[
    frameworkbox,
    fill=pal1,
    right=0.45cm of mllm
] (captionlayer) {%
    \textbf{Descriptive grounding}\\[1pt]
    \emph{Caption}\\[1pt]
    Visible objects, people, and physical setting
};

\node[
    frameworkbox,
    fill=sentNeu,
    right=0.25cm of captionlayer
] (taglayer) {%
    \textbf{Intermediate semantic layer}\\[1pt]
    \emph{Perception tags}\\[1pt]
    Structured attributes assigned to the scene
};

\node[
    frameworkbox,
    fill=sentSPos,
    right=0.25cm of taglayer
] (justlayer) {%
    \textbf{Interpretive framing}\\[1pt]
    \emph{Justification}\\[1pt]
    Evaluation, contextualization, and meaning
};

\draw[flowarrow]
    (input.east) -- (mllm.west);


\coordinate (buslevel) at
    ([yshift=0.52cm]taglayer.north);

\coordinate (modelout) at
    ([xshift=0.22cm]mllm.east);

\coordinate (busleft) at
    (modelout |- buslevel);

\coordinate (captiondrop) at
    (captionlayer.north |- buslevel);

\coordinate (tagdrop) at
    (taglayer.north |- buslevel);

\coordinate (justdrop) at
    (justlayer.north |- buslevel);

\draw[
    flowline,
    rounded corners=2pt
]
    (mllm.east)
    -- (modelout)
    -- (busleft);

\draw[flowline]
    (busleft) -- (justdrop);

\draw[flowarrow]
    (captiondrop) -- (captionlayer.north);

\draw[flowarrow]
    (tagdrop) -- (taglayer.north);

\draw[flowarrow]
    (justdrop) -- (justlayer.north);

\draw[
    ->,
    thick,
    gray!55
]
([yshift=-0.27cm]captionlayer.south west)
--
node[
    below,
    font=\scriptsize\itshape,
    text=black!70
]
{Increasing interpretive abstraction}
([yshift=-0.27cm]justlayer.south east);

\end{scope}


\begin{scope}[yshift=-3.65cm]


\node[
    examplebox,
    fill=pal1
] (captions) at (2.00,0.12) {%
    \textbf{\underline{Descriptive grounding}: \textit{Captions}}\\[2pt]
    \textbf{P1 -- Female, Low Income, Conservative, Pragmatic:}
    ``A tall brick building with a stone base, surrounded by greenery
    and a clear sky.''\\[1pt]
    \textbf{P2 -- Female, Low Income, Progressive, Empathetic:}
    ``A university campus with tall brick buildings, a tree, and a
    clear sky with people near the entrance.''\\[1pt]
    \textbf{P3 -- Male, Low Income, Conservative, Analytical:}
    ``A tall brick building with a stone base, surrounded by trees and
    a clear sky.''
};


\node[
    examplebox,
    fill=sentNeu,
    below=0.08cm of captions
] (perceptions) {%
    \textbf{\underline{Intermediate semantic layer}:
    \textit{Perception tags}}\\[2pt]
    \textbf{P1:}
    Urban environment; Cleanliness; Everyday environment;
    Old building; Well-lit environment.\\[1pt]
    \textbf{P2:}
    Pleasant environment; Everyday environment.\\[1pt]
    \textbf{P3:}
    Old building; Everyday environment; Cleanliness.
};


\node[
    examplebox,
    fill=sentSPos,
    below=0.08cm of perceptions
] (justifications) {%
    \textbf{\underline{Interpretive framing}:
    \textit{Justifications}}\\[2pt]
    \textbf{P1:}
    ``A clean, well-maintained city scene with orderly architecture,
    fitting a conservative view of practical urban upkeep.''\\[1pt]
    \textbf{P2:}
    ``This campus feels welcoming and well-maintained, aligning with
    values of accessible education.''\\[1pt]
    \textbf{P3:}
    ``Looks like a stable, traditional urban setting with maintained
    buildings and greenery.''
};

\coordinate (annotationmid) at
    ($(captions.north)!0.5!(justifications.south)$);

\coordinate (imageeast) at
    ([xshift=-0.60cm]captions.west);

\node[
    inner sep=0pt,
    anchor=east
] (img) at (imageeast |- annotationmid) {%
    \includegraphics[
        width=4.35cm,
        height=4.75cm,
        keepaspectratio
    ]{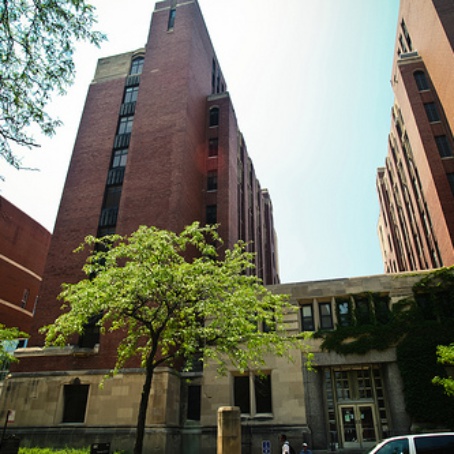}
};

\node[
    anchor=south west,
    font=\small\bfseries
] at ([xshift=-4.78cm,yshift=0.18cm]captions.north west)
{(b) Persona-conditioned annotation example};

\draw[flowarrow]
    (img.east |- captions.west)
    -- (captions.west);

\draw[flowarrow]
    (img.east |- perceptions.west)
    -- (perceptions.west);

\draw[flowarrow]
    (img.east |- justifications.west)
    -- (justifications.west);

\end{scope}

\end{tikzpicture}\vspace{-0.3cm}
\caption{
Functional framework and persona-conditioned example. Panel (a) distinguishes descriptive grounding, intermediate semantic selection, and interpretive framing. Panel (b) groups three outputs for the same image by type: captions converge on visible content, whereas perception tags and justifications vary more in semantic emphasis and evaluative framing.
}
\label{fig:annotation_example}
\end{figure*}

In our setting, a model annotates each urban image with a sentiment label on a five-point scale (from Negative to Positive), a caption, a set of perception tags, and a justification. We organize these outputs into three functional layers (Figure~\ref{fig:annotation_example}): \emph{descriptive grounding}, represented by captions of visible content; an \emph{intermediate semantic layer}, represented by perception tags; and \emph{interpretive framing}, represented by justifications that explain the assigned sentiment through the evaluation and contextualization of the scene. The three layers leave different room for a persona to shape the output: captions are anchored in visual evidence shared by all observers, perception tags require selecting labels from a controlled vocabulary, and justifications are open-ended and explicitly evaluative. If persona conditioning acts mainly on interpretation, its effects should be more apparent in perception tags and justifications than in captions. Persona profiles may therefore produce similar descriptions of the same visual evidence while emphasizing different interpretations or evaluative aspects. Our central finding is that persona conditioning primarily affects interpretive framing, while descriptive grounding remains comparatively stable.

We examine this distinction using approximately $60{,}000$ persona-conditioned annotations from each of two MLLMs: Qwen3-VL and Gemma4 (Section~\ref{sec:method}). For each model, $1{,}200$ persona-conditioned agent instances annotate the same urban scenes under matched prompts, personas, and decoding settings, alongside no-persona baselines.

Our contributions are fourfold: (i) we introduce a functional framework distinguishing descriptive grounding, an intermediate semantic layer, and interpretive framing; (ii) we show that persona-associated variation is substantially stronger in interpretive than in descriptive language, supported by paired image-level comparisons of captions and justifications on a common cosine-similarity scale; (iii) we identify economic status as the most consistent source of interpretive variation, with political orientation, personality, and gender exhibiting smaller or model-dependent patterns; and (iv) we assess cross-model consistency and test whether the economic status pattern persists under deterministic decoding. Code, prompts,
configurations, and processed outputs are publicly available\footnote{\url{https://minds-lab-utfpr.github.io/Persona-Interpretive-Analysis}}.

\section{Related Work}\label{sec:related}

Prior work shows that LLMs can reproduce stylized experimental findings and simulate decision-making processes \cite{aherICML2023,Apostolos2024}, supporting applications in surveys, experiments, and content analysis \cite{Bail2024PNAS}. However, important limitations remain, including bias, limited validity, and difficulties in representing human diversity \cite{wang_zou_yan_guo_sun_xiao_zhang_2024,wang2025large}.

A related line of research investigates LLMs as synthetic respondents and simulated populations. Studies show that LLMs can approximate human-like responses and support large-scale agent simulations \cite{Argyle_Busby_Fulda_Gubler_Rytting_Wingate_2023,park2024generativeagentsimulations1000}, while also revealing limitations in modeling belief structures and reasoning processes \cite{li2025simulating,barrie_cerina_2026}. These findings suggest that evaluation should extend beyond predictive accuracy to include consistency and interpretability.

Persona prompting plays a central role in this context. Conditioning models on demographic or psychological attributes can influence outputs, although effects are often limited or task-dependent \cite{hu-collier-2024-quantifying,beck-etal-2024-sensitivity}. Most related work evaluates persona effects on structured outputs — labels, ratings, or choices — leaving open the question of whether and how personas shape the language of open-ended responses \cite{lutz-etal-2025-prompt,malik-etal-2025-llms}.

Our work also connects to research on multimodal LLMs and urban perception. Prior studies show that urban image perception is subjective and socially contextualized, with perceptions of safety, disorder, beauty, and wealth varying systematically across observers and environments \cite{he2026urbanfeela,balsa2026urbanexperiences,perceptSentPaper,Xu_Pan_Liu_Wang_2025,10.1145/3385186}. Recent work further shows that urban visual perception and its relationship with sentiment vary across socioeconomic contexts, with differences observed between lower- and higher-income communities \cite{shen26}. Multimodal LLMs can partially capture some of these perceptual dimensions \cite{wu2025cycling,dasilva2025multimodalllmssentiment}. At the same time, recent work reports strong within-persona consistency but limited cross-persona variation in outputs such as sentiment labels \cite{urbcom26-neemias}, raising the question of whether persona effects emerge more clearly in interpretive language than in structured outputs.

We address this gap directly, analyzing captions, justifications, and perception tags generated by MLLMs to examine how persona effects emerge in descriptive grounding and interpretive framing. The key question is whether persona conditioning produces structured variation in how scenes are interpreted, even when descriptive content converges. 

\section{Data and Methodology}\label{sec:method}

\subsection{Dataset and Conditions}

We analyze two datasets. The first is a publicly available \emph{Qwen3-VL} 
annotation corpus~\cite{urbcom26-neemias}, comprising 60,208 records generated under persona-conditioned and no-persona settings for 50 urban scene images. Each record contains four outputs: (i) a predicted sentiment label on a five-point scale (Negative, Slightly Negative, Neutral, Slightly Positive, Positive), (ii) a caption describing the image, (iii) one or more perception tags  (the prompt requests one to five) selected from a controlled vocabulary of 593 labels, and (iv) a justification explaining the assigned sentiment. Captions and justifications contain \(16.9 \pm 5.0\) and \(18.0 \pm 4.9\) words on average, respectively. Both persona-conditioned and no-persona justifications are written as full prose explanations rather than simple label rationales, enabling linguistic analysis beyond sentiment labels. Figure~\ref{fig:annotation_example}b illustrates one annotation record per persona for the same urban image.

All annotations were generated under matched decoding settings (\texttt{temperature=0.1}, \texttt{seed=42}), following the configuration documented in the accompanying code repository and prior dataset description~\cite{urbcom26-neemias}. We conduct a targeted decoding-sensitivity ablation at temperature $T=0$ for economic status (Appendix~\ref{app:temperature0}).  We use the publicly available dataset and construct a second dataset with \emph{Gemma4} 
using the same images, prompts, persona profiles, and decoding settings. The prompt templates are given in Appendix~\ref{app:prompts}. The two models are served differently: Qwen3-VL is \texttt{Qwen3-VL-8B} run through Ollama (tag \texttt{qwen3-vl:8b}, Q4\_K\_M quantization), whereas Gemma4 is \texttt{google/gemma-4-E4B-it} run through Hugging Face \texttt{transformers} with bf16 weights. This matched design reduces, though does not eliminate, differences attributable to the experimental configuration, allowing us to compare persona-conditioned patterns across the two models. For each model, we consider three annotation conditions: persona-conditioned generation, no-persona generation with extended reasoning, and no-persona generation without extended reasoning.

\begin{itemize}
\item \textbf{Persona.}
The main corpus encompasses annotations produced using $|P|=24$ persona profiles obtained from the Cartesian product of $|D|=4$  demographic and personality attributes: \emph{gender} (Male, Female), \emph{economic status} (Low, High), \emph{political orientation} (Conservative, Progressive), and \emph{personality} (Pragmatic, Empathetic, Analytical). Each profile is instantiated as $|A|=50$ persona-conditioned agent instances ($24 \times 50 = 1{,}200$ in total), and each instance annotates all ($|I|=50$) images using persona-conditioned prompts with extended reasoning enabled (\texttt{think=True}). We use the term \emph{persona-conditioned agent instance} to denote a generation replicate under a fixed persona profile and decoding configuration. These instances do not interact, maintain memory, or act autonomously; each annotation is generated through a separate inference call under the same decoding configuration. The full list of profiles is shown along the rows of Figure~\ref{fig:ic_caption} (Appendix~\ref{app:InterProfSim}).

\item \textbf{No-Persona.}  A no-persona setting, in which the same model annotates the same images using a neutral observer prompt, includes two variants: with extended reasoning  (\textbf{NPT}, \texttt{think=True}) and without  (\textbf{NPNoT}, \texttt{think=False}). The two variants produce five annotations per image ($250$ each), providing complementary baselines for the persona-generated corpus.

\end{itemize}

Table \ref{tab:annotation_counts} summarizes all annotations per model and experiment condition.

\begin{table}[htb] \centering \scriptsize \begin{tabular}{p{1.2cm}p{0.8cm}p{0.5cm}p{0.8cm}lp{1cm}} \toprule
& \multicolumn{3}{c}{Experiment Condition} & & \multicolumn{1}{c}{Total} \\
Model & Persona & NPT &  NPNoT  & Missing & \multicolumn{1}{c}{Observed}  \\
\midrule
Qwen3-VL & 60,000 & 250 & 250 & 292 & 60,208 \\
Gemma4   & 60,000 & 250 & 250 & 1 & 60,499 \\
\bottomrule
\end{tabular}\vspace{-0.3cm}
\caption{Expected and observed numbers of annotations by model. The expected total is 60,500 annotations per model. Missing records correspond to failed generations relative to this total. No missing records occurred in the NPT or NPNoT cases.}
\label{tab:annotation_counts}
\end{table}

In the following subsections, we evaluate within-persona stability and cross-persona divergence separately for captions, justifications, and perception tags. For each matrix, we compute pairwise similarities between agent annotations, comparing agents that share the same persona profile or attribute level with those that do not, depending on the analysis.

We summarize diagonal separation using the Diagonal Strength Index (DSI) to quantify how much stronger the diagonal elements of a matrix (representing similarities between agents from the same persona profiles) are compared with the off-diagonal elements (representing agents from different persona profiles): $DSI=\frac{\mu_D-\mu_O}{\mu_D},$ where $\mu_D$ denotes the mean of the diagonal elements; $\mu_O$ is the mean of the off-diagonal elements.

\subsection{Evaluating Within-Persona Stability}\label{secWithPerStability}

 We first assess the internal stability of agents sharing the same persona, i.e., the reproducibility of generated outputs when all demographic and personality attributes are held constant. We evaluate stability across three output types:
 
\begin{itemize}
  \item \textbf{Captions (descriptive grounding) and justifications (interpretive framing):} Captions and justifications are encoded using the Sentence-BERT model \texttt{all-MiniLM-L6-v2}, and semantic similarity is computed as the cosine similarity between the resulting sentence embeddings.

  \item \textbf{Perception tags (intermediate semantic layer):}
Each annotation contains a set of perception tags drawn from a controlled vocabulary. These tags form an intermediate semantic layer between descriptive grounding and interpretive framing, capturing structured aspects of scene interpretation. Because the tags are discrete labels rather than unconstrained text, we measure exact set overlap using Jaccard similarity, defined as the intersection over the union of the two tag sets. Records with an empty tag list (13 in Qwen3-VL, none in Gemma4) are treated as missing and excluded from perception pairs rather than scored as zero overlap.

\end{itemize}

For each type (captions, justifications, or perception tags), we compute a distribution of within-persona similarities and summarize it for each profile using the mean. Higher values indicate greater convergence among outputs generated under the same persona profile. Topic modeling was retained only for justifications because captions showed high cross-agent similarity and insufficient semantic variation to support meaningful thematic discrimination.

\subsection{Evaluating Cross-Persona Divergence}

We evaluate whether outputs differ systematically across four persona dimensions $D$: gender, economic status, political orientation, and personality. For each image $i\in I$ and dimension $d\in D$, annotation pairs are partitioned into:

\begin{itemize}
    \item \textbf{Within-group pairs $\odot$:} agents sharing the same attribute level (e.g., two high-income agents), regardless of the remaining persona dimensions.

    \item \textbf{Cross-group pairs $\otimes$:} agents with different attribute levels (e.g., high-income vs.\ low-income), regardless of the remaining persona dimensions. For personality, which has three levels, this set pools the three possible level contrasts.
\end{itemize}

As a sensitivity analysis, we also compare matched profiles that differ only on the target dimension (Appendix \ref{app:matched_factorial}).

We compare groups across complementary output types:

\begin{itemize}
    \item \textbf{Captions (descriptive grounding) and justifications
    (interpretive framing):}
    We compute pairwise cosine similarity between sentence embeddings, as described in Section 3.2.

    \item \textbf{Perception tags (intermediate semantic layer):}
    We compute Jaccard similarity between tag sets, as described in
    Section~\ref{secWithPerStability}.

    \item \textbf{Topic distributions (interpretive framing):}
    For each image and persona dimension, we construct topic distributions for each attribute level after excluding unassigned outlier documents (i.e., BERTopic topic $-1$). We compute Jensen--Shannon divergence (JSD) between attribute-level pairs---one pair for each binary dimension and three pairs for personality---and average the values across pairs and images. Significance is assessed using $1{,}000$ within-image label permutations that preserve group sizes. Heatmaps show the most frequent topic proportions across personas.
\end{itemize}

For captions, justifications, and perception tags, we take the image as the unit of analysis. For each image $i$ and persona dimension $d$, we compute the mean similarity among annotation pairs sharing the same attribute level, $\bar{s}_{i,d,\odot}$ and among pairs with
different attribute levels, $\bar{s}_{i,d,\otimes}$. The paired image-level difference is set as $\Delta_{i,d}=\bar{s}_{i,d,\odot}-\bar{s}_{i,d,\otimes}$. Positive values indicate that annotations sharing the target attribute level are, on average, more similar than annotations with different attribute levels. We denote $\Delta_d=\frac{1}{|I|}\sum_{i \in I}\Delta_{i,d}$ as the mean difference across images.

Because each image contributes a within-group and a cross-group mean, we conduct a separate paired analysis for each model, output type, and persona dimension. For image $i$, the paired observations are $\bar{s}_{i,d,\odot}$ and $\bar{s}_{i,d,\otimes}$. Equivalently, we apply a two-sided Wilcoxon signed-rank test to the 50 image-level differences $\{\Delta_{i,d}: i \in I\}$, testing whether their location is zero. We report $\Delta_d$, the mean difference across images, with a 95\% bias-corrected and accelerated (BCa) bootstrap confidence interval obtained by resampling images. Benjamini--Hochberg correction is applied across the 12 tests within each model. Because the signed-rank test and $\Delta_d$ summarize different aspects of the distribution, interpretation emphasizes $\Delta_d$ and its confidence interval, with statistical significance as complementary evidence. Full statistics, effect sizes, sensitivity analyses, and complete-case results appear in Appendix~\ref{app:stats}.

To compare descriptive and interpretive text directly, we conduct a second paired analysis. For each model and persona dimension $d$, we apply the same procedure to the 50 image-level contrasts $\Delta^{\mathrm{just}}_{i,d}-\Delta^{\mathrm{cap}}_{i,d}$, pairing justification and caption differences from the same image. Benjamini--Hochberg correction is applied across the four persona dimensions within each model.  Perception-tag differences use Jaccard similarity and are therefore analyzed separately.

For topic distributions, a larger Jensen--Shannon divergence indicates greater variation in the themes emphasized across persona groups.

\section{Results}
\label{sec:results}

\subsection{Persona Effects from a Profile Perspective}
\label{sec:results-wihin} 

For each profile pair $(p_x,p_y)$, similarity is computed within each image and then averaged across shared images, yielding a $24 \times 24$ inter-profile similarity matrix for each model (Appendix \ref{app:InterProfSim}).

\begin{table}[t]
\scriptsize
\centering
\begin{tabular}{lcccl}
  \toprule
         & \multicolumn{3}{c}{Diagonal Strength Index (\%)}                                                    &  \\
Model    & Captions & \multicolumn{1}{l}{Justifications} & \multicolumn{1}{l}{Perception} &  \\
  \midrule
Qwen3-VL & 2.3      & 14.9                               & 11.7                            &  \\
Gemma4   & 3.7      & 30.0                               & 25.1                            &  \\
 \bottomrule
\end{tabular}\vspace{-0.28cm}
\caption{DSI (\%) for both models and output types.}
\label{tab:DSI}
\end{table}

Table \ref{tab:DSI} shows DSI (\% values) for both models and the three types of outputs.  It reveals a consistent pattern across the models. Caption generation exhibits low DSI values (2.3--3.7\%), indicating weak profile-associated separation in descriptive content. In contrast, DSI is higher for perception tags (11.7--25.1\%) and highest for justifications (14.9--30.0\%), indicating stronger profile-associated structure in interpretive outputs. Across all three output types, Gemma4 produces higher DSI values than Qwen3-VL, indicating stronger separation between within-profile and cross-profile similarity. In both models, captions show the lowest DSI, whereas perception tags and justifications show stronger profile-associated separation.

Table~\ref{tab:cross_model_corr} summarizes cross-model agreement between the resulting matrices. It shows strong agreement across all output types. Captions show the highest cross-model agreement, whereas justifications show the lowest.

\begin{table}[t]
    \centering
    \scriptsize
    \begin{tabular}{l|c|c}
        \toprule
        Type & Pearson $r$ & Spearman $\rho$\\
        \midrule
        Captions (cosine)          & $0.89$ & $0.87$ \\
        Justifications (cosine)     & $0.80$ & $0.78$  \\ 
        Perception tags (Jaccard)   & $0.88$ & $0.83$  \\
        \bottomrule
    \end{tabular}\vspace{-0.3cm}
\caption{
Cross-model agreement between Qwen3-VL and Gemma4, measured by correlating the upper triangles of their image-conditioned inter-profile similarity matrices. Higher correlations indicate more similar profile-pair rankings across models.}
    \label{tab:cross_model_corr}
\end{table}

\subsection{Persona Effects from a Dimensional Perspective}
\label{sec:concordance}

We now examine how similarity is associated with individual persona dimensions across three output types: captions (\emph{descriptive grounding}), justifications (\emph{interpretive framing}), and perception tags (\emph{intermediate semantic layer}). Building on the convergence observed in Section~\ref{sec:results-wihin}, this analysis asks whether semantic similarity is systematically structured by persona attributes.

\begin{figure}[t]
\centering
\subfloat[Qwen3-VL]{%
    \includegraphics[
        width=.98\columnwidth
    ]{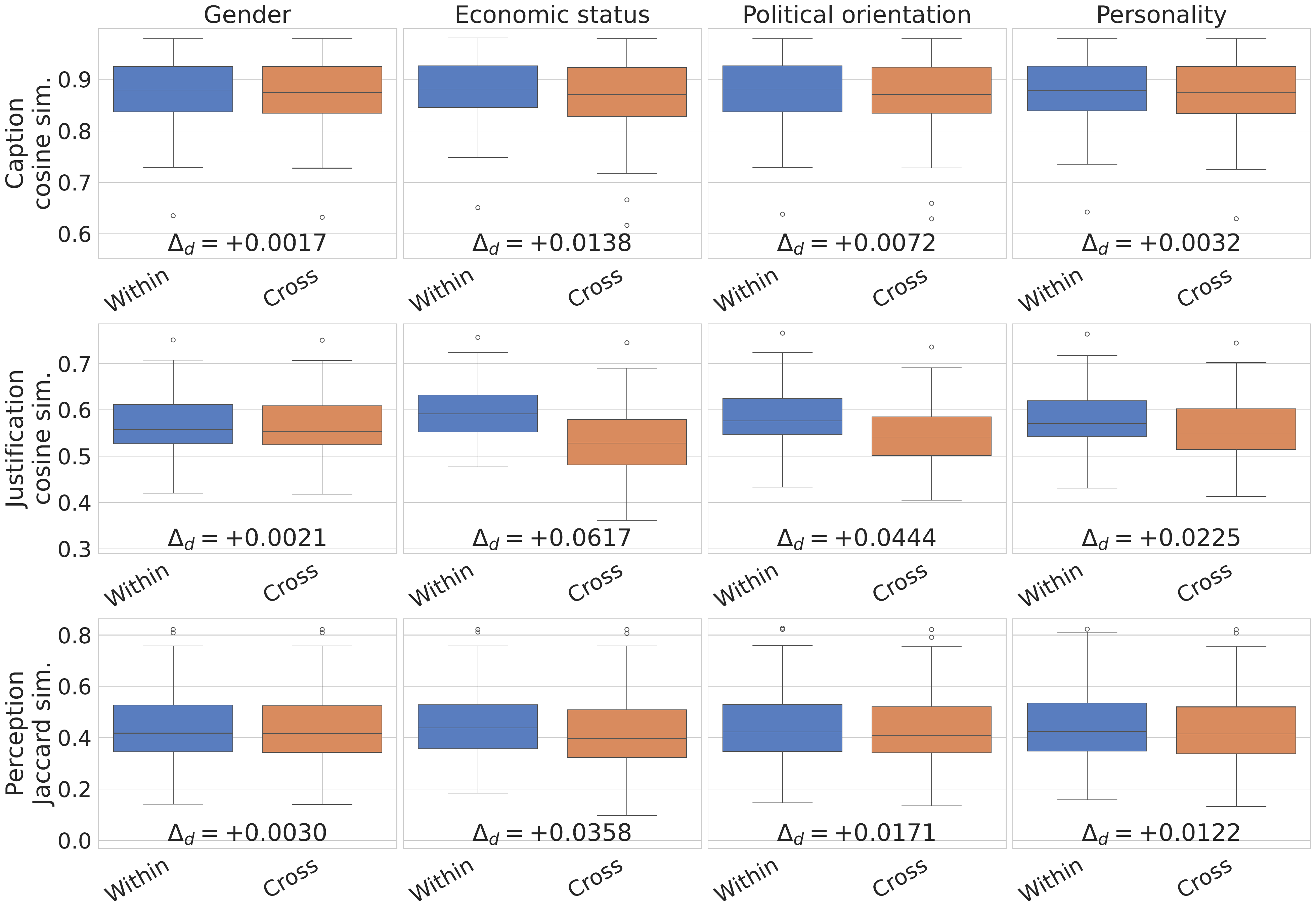}
}\\[-1pt]
\subfloat[Gemma4]{%
    \includegraphics[
        width=.98\columnwidth
    ]{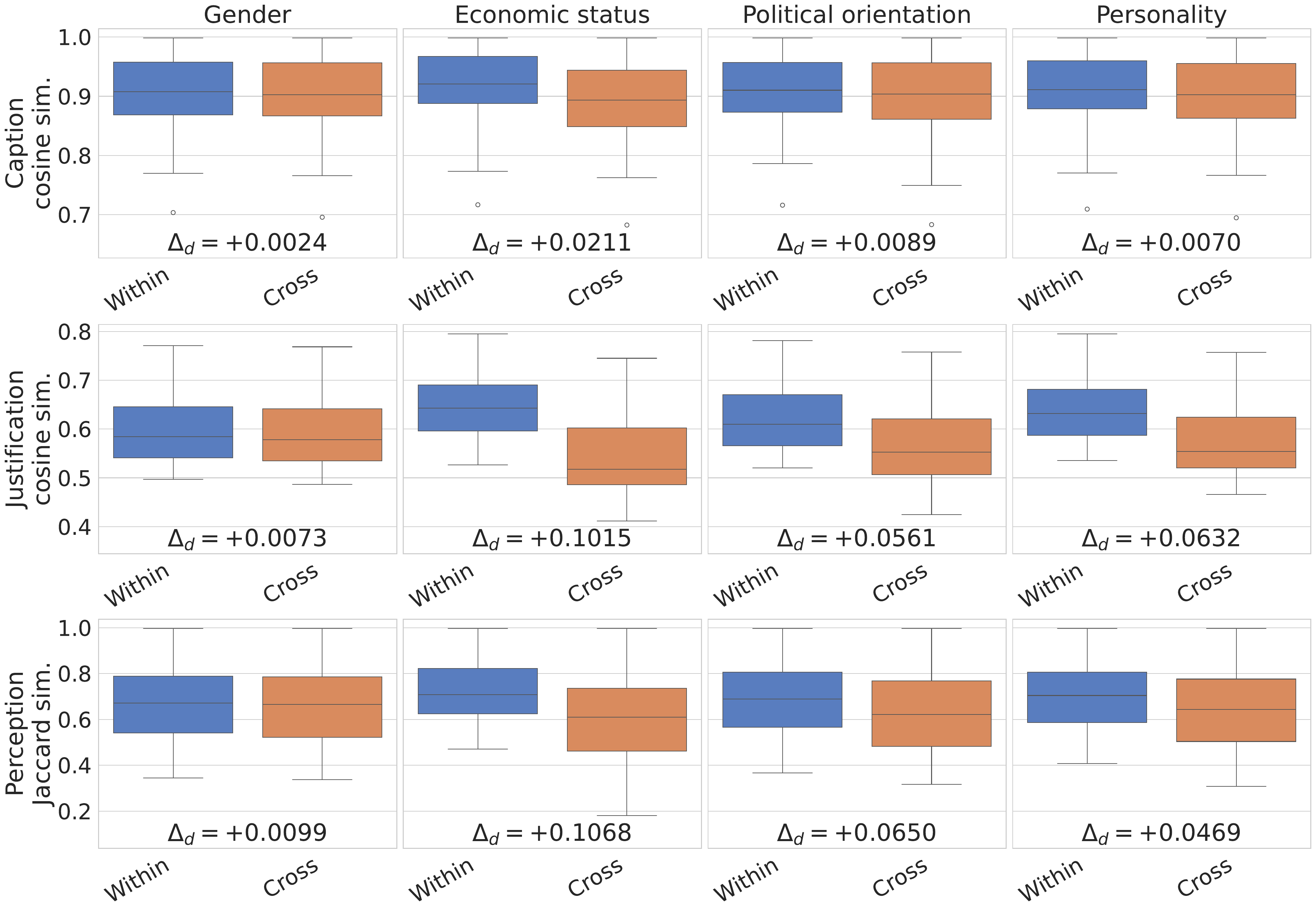}
}
\caption{Within-group ($\odot$) \emph{vs.}\ cross-group ($\otimes$) similarity for captions (top row), justifications (middle row), and perception tags (bottom row) across persona dimensions. The mean image-level difference $\Delta_d$ is shown beneath each pair of boxplots. Captions and justifications use cosine similarity, whereas perception tags use Jaccard similarity. Full confidence intervals, effect sizes, and adjusted tests are reported in Appendix~\ref{app:stats}.}
\label{fig:within_cross}
\end{figure}

Figure~\ref{fig:within_cross} compares within-group and cross-group similarity across four persona dimensions. Captions and justifications are measured using cosine similarity, whereas perception tags are measured using Jaccard similarity.

For captions, within-group similarity exceeds cross-group similarity for every persona dimension in both models, but the differences are small: $\Delta_d$ ranges from $+0.002$  for gender in Qwen3-VL to $+0.021$ for economic status in Gemma4, while mean caption similarities remain approximately $0.87$--$0.91$. Descriptive grounding, therefore, exhibits detectable but limited attribute-associated variation, with captions remaining highly similar across persona groups.

Justifications show larger attribute-associated differences. Economic status produces the largest difference in both models (Qwen3-VL: $\Delta_d=+0.062$; Gemma4: $\Delta_d=+0.101$), followed by political orientation in Qwen3-VL ($\Delta_d=+0.044$). In Gemma4, personality ($\Delta_d=+0.063$) and political orientation ($\Delta_d=+0.056$) are also prominent. Gender produces the smallest justification difference in both models (Qwen3-VL: $\Delta_d=+0.002$; Gemma4: $\Delta_d=+0.007$).

Direct paired contrasts on the common cosine scale ($\Delta^{\mathrm{just}}_{i,d}-\Delta^{\mathrm{cap}}_{i,d}$) show that justification differences exceed caption differences for economic status, political orientation, and personality. The mean image-level contrasts ($\Delta^{\mathrm{just}}_{d}-\Delta^{\mathrm{cap}}_{d}$)  range from $+0.019$ to $+0.048$ in Qwen3-VL and from $+0.047$ to $+0.080$ in Gemma4. Gender again produces the smallest contrast; for Qwen3-VL, its mean contrast is $+0.0004$, and the BCa confidence interval includes zero. Full paired results are reported in
Appendix~\ref{app:stats}.

For perception tags, economic status produces the largest within--cross difference in both models (Qwen3-VL: $\Delta_d=+0.036$; Gemma4: $\Delta_d=+0.107$), whereas gender produces the smallest ($+0.003$ and $+0.010$, respectively). Political orientation and personality fall between these extremes. Gemma4 shows larger Jaccard differences than Qwen3-VL across all dimensions; these values are interpreted separately from the cosine-based ones.

These results locate persona-associated variation primarily in interpretive language. Justifications exhibit substantially larger within--cross differences than captions, particularly for economic status and political orientation, while caption differences remain small. Perception tags also show greater within-group than cross-group similarity across all dimensions.

Within each output type, Gemma4 generally exhibits larger within--cross differences than Qwen3-VL, although economic status remains the largest dimension and gender the smallest in both models. This pattern reinforces the central distinction of the paper: \emph{persona conditioning affects interpretive framing more strongly than descriptive grounding}. The economic status pattern also persists under deterministic decoding ($T=0$; Appendix~\ref{app:temperature0}).

A matched-factorial sensitivity analysis that holds the remaining persona attributes fixed preserves the broad dimensional ordering (Appendix~\ref{app:matched_factorial}), while a paired justification-minus-caption contrast within the matched design confirms larger differences for all four attributes (Appendix~\ref{app:matched_factorial_contrast}).

Additional comparisons with no-persona generations show similarly strong descriptive convergence but greater variability in justificatory language; full results are reported in Appendix~\ref{app:nopersona}.

\subsection{Persona Effects on Topic Structure}
\label{sec:topic-results}

We further explore interpretive framing through the thematic content of justifications. We focus on justifications because captions yielded semantically uniform and unstable BERTopic results, with insufficient variation to produce differentiated topics. The ten most prevalent justification topics include evaluative themes related to natural beauty, development, accident response, tranquility, decay, and neglect. Both models are analyzed using the same BERTopic pipeline. Topic nicknames are manually assigned by the authors after inspecting the highest-weight c-TF-IDF terms and BERTopic's representative documents for each topic.

To quantify how strongly persona dimensions structure thematic content, Table~\ref{tab:topic_jsd} reports the between-group JSD of justification-topic distributions. Within both models, economic status produces the largest divergence and gender the smallest. Political orientation is the second-largest dimension for Qwen3-VL, whereas personality and political orientation have similar magnitudes in Gemma4. All dimensions reach significance under the within-image permutation test ($p_{\mathrm{emp}}\leq0.001$), but substantive interpretation emphasizes JSD magnitude rather than significance alone.

\begin{table}[t]
    \centering
    \scriptsize
    \begin{tabular}{lcc}
        \toprule
        Persona dimension & Qwen3-VL & Gemma4 \\
        \midrule
        Economic status       & $0.311$ & $0.718$ \\
        Political orientation & $0.281$ & $0.557$ \\
        Personality           & $0.095$ & $0.579$ \\
        Gender                & $0.026$ & $0.172$ \\
        \bottomrule
    \end{tabular}
    \vspace{-0.25cm}
    \caption{
    Between-group Jensen--Shannon divergence of justification-topic distributions, averaged over group pairs and images. Higher values indicate greater thematic divergence within a model. Absolute values are not directly comparable across models because their BERTopic models were fitted independently.
    }
    \label{tab:topic_jsd}
\end{table}

\begin{figure}[t]
    \centering
    \subfloat[Qwen3-VL]{%
        \includegraphics[
            width=1.01\columnwidth
        ]{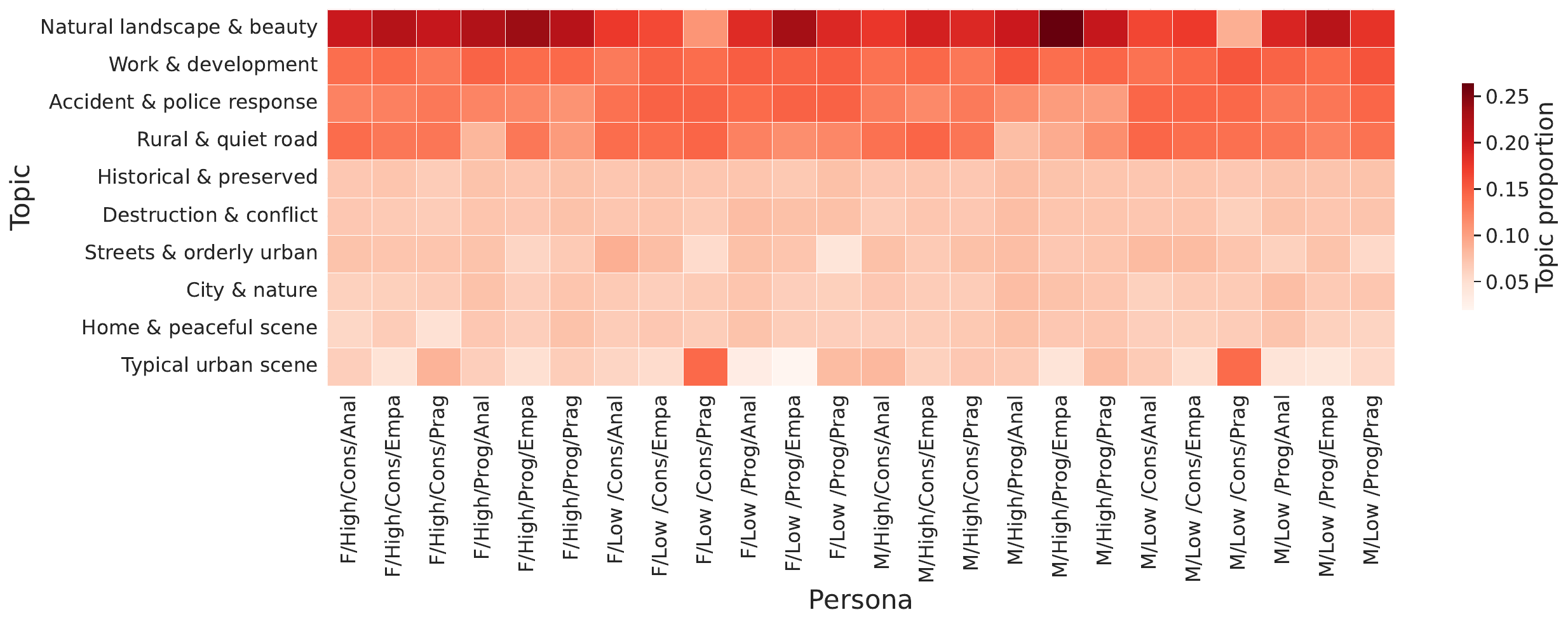}
    }\\[-1pt]
    \subfloat[Gemma4]{%
        \includegraphics[
            width=1.01\columnwidth
        ]{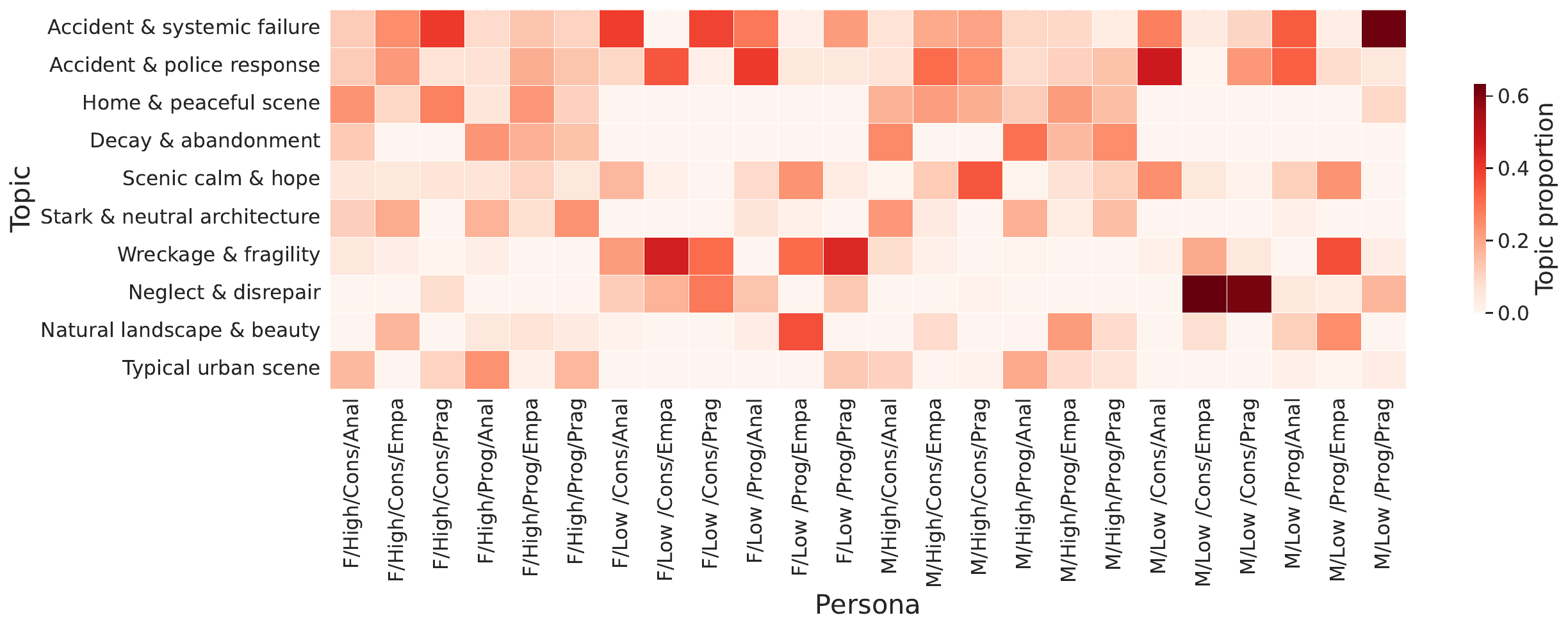}
    }
    \vspace{-0.25cm}
    \caption{
    Column-normalized proportions of the ten most prevalent assigned justification topics across persona profiles. BERTopic was fitted independently for Qwen3-VL and Gemma4; topic labels, proportions, and color intensities should therefore be interpreted within each panel rather than compared directly across models.
    }
    \label{fig:topic_persona_heatmap}
\end{figure}

Figure~\ref{fig:topic_persona_heatmap} 
shows differences in a profile-level view. For Qwen3-VL, most visible contrasts occur across economic status and political orientation profiles, particularly for topics related to neglect, development, order, and natural beauty. Gender and personality profiles appear comparatively uniform, consistent with their smaller JSD values.

Some Qwen3-VL topics, including \emph{Historical \& preserved} and \emph{City \& nature}, remain relatively stable across profiles. Others, such as \emph{Typical urban scene} and \emph{Natural landscape \& beauty}, vary more visibly. This suggests that persona conditioning changes which evaluative aspects of the same scene become thematically salient.

Gemma4 exhibits economic status variation, although the direction is topic-specific rather than uniformly negative or positive. In its heatmap, accident- and neglect--related topics appear more often in some low-income profiles, and peaceful, natural, and architectural themes in selected high-income profiles. These are profile-level patterns over the ten displayed topics rather than a tested directional effect of economic status.

As shown in Figure~\ref{fig:topic_sentiment}, Appendix~\ref{app:TopicsAnalysis}, both models show consistent alignment between justification topics and sentiment, with accident, conflict, neglect, and decay topics strongly associated with negative sentiment, and nature, beauty, and peaceful scenes with positive sentiment. In contrast, everyday urban infrastructure and typical city scenes are predominantly neutral or mixed, reflecting the ambiguity of ordinary urban environments.

These patterns require cautious interpretation. BERTopic assigned topic $-1$ to $15\%$ of Qwen3-VL justifications and $57\%$ of Gemma4 justifications. Because the analysis conditions on assigned topics, and the displayed topics cover different proportions of the two corpora, absolute JSD values and heatmap contrasts should not be compared directly across models.

The sharper Gemma4 contrasts therefore do not by themselves indicate a stronger persona effect. The more reliable cross-model result is the within-model ordering of dimensions: economic status produces the largest thematic divergence and gender the smallest.

Overall, the topic analysis complements the similarity results: persona conditioning changes the thematic emphasis of justifications, supporting the distinction between shared descriptive grounding and variable interpretive framing.

\section{Conclusion}\label{sec:conclusion}

This work examined how persona prompting shapes language generated by Qwen3-VL and Gemma4 for urban perception. We introduced a functional framework distinguishing descriptive grounding, an intermediate semantic layer, and interpretive framing.

Across roughly $60{,}000$ persona-conditioned annotations per model, captions remained highly similar across profiles, whereas justifications showed substantially greater variation, particularly for economic status, political orientation, and personality. Economic status produced the largest differences and gender the smallest, an ordering broadly preserved by the matched-factorial analysis. Perception tags likewise showed greater within-group than cross-group similarity across attributes.

The prominence of economic status is consistent with recent urban-perception research reporting socioeconomic differences in visual perception and its relationship with sentiment \cite{shen26}. This parallel does not establish that persona-conditioned outputs reproduce human income-group differences, but highlights model sensitivity to economic status prompting.

Gemma4 generally exhibited larger persona-associated differences than
Qwen3-VL, although their magnitudes may also reflect differences in
serving stack and precision. Nevertheless, the shared dimensional ordering and strong cross-model agreement indicate a consistent qualitative pattern whose magnitude varies across the evaluated configurations.

Overall, structured labels and descriptive outputs may understate persona-conditioned variation, which emerges most clearly in open-ended explanations. Persona-conditioned MLLMs should therefore be treated as generators of prompted interpretive perspectives rather than simulations of human demographic groups. Future work should validate these patterns against demographically matched human annotations and across additional models, languages, and visual domains.

\section*{Limitations and Ethics Considerations}
 
Our factorial design represents personas through a controlled set of demographic and personality attributes. Although this enables attribute-level comparisons, the profiles are prompting conditions rather than comprehensive representations of individual identities. We therefore interpret the observed differences as persona-conditioned model behavior, not as estimates of how corresponding human groups perceive urban scenes. Establishing such correspondence would require demographically matched human annotations.

Similarly, the 50 instances generated for each persona profile should not be interpreted as 50 independent or autonomous agents. They are repeated model inferences under the same persona-conditioned configuration and are used to characterize variability across generations. Differences among them may therefore reflect decoding stochasticity rather than systematic instance-level variation. Future work could introduce persistent identities or state when modeling distinct simulated individuals.

Using the same 50 images for both models enables a controlled comparison but limits the generality of the findings, which are specific to the evaluated models, prompts, personas, and image set. Future work should assess their robustness across additional configurations, models, and datasets.

Several design choices qualify our comparisons. First, the prompt asks for captions to be written independently of the persona, whereas justifications are produced in the persona's own voice (Appendix~\ref{app:prompts}). The observed gap between descriptive and interpretive outputs may therefore partly reflect this task design, although captions still exhibit small but systematic attribute-associated differences. Second, justifications may repeat or paraphrase persona cues from the prompt, so some within-group similarity may reflect shared lexical content rather than interpretive framing alone. Future work should test this possibility by masking persona-related terms before recomputing $\Delta_d$ and by controlling for output length. Third, the models differ in architecture, serving stack, and weight precision; cross-model differences should therefore be interpreted as differences between the evaluated configurations rather than architecture alone. Finally, the topic analysis is exploratory: BERTopic leaves $57\%$ of Gemma4 justifications unassigned. A shared topic model and directional analyses of topic and sentiment distributions by attribute level remain important extensions.

Synthetic personas may reproduce biased or stereotypical associations. To support scrutiny of these risks, we release the prompts, configurations, and analysis code for replication and auditing.

\section*{Acknowledgments}

This study was supported in part by the National Council for Scientific and Technological Development - CNPq (processes 441444/2023-7, 403646/2026-0, and 444724/2024-9) and INCT TILD-IAR (proc. 408490/2024-1). This research was also supported in part by a CIFAR AI Catalyst Grant, ``Towards Socially Grounded AI Safety: Integrating Causal and Institutional Reasoning in Language Models'' (PI: Matt Ratto; Co-PI Zhijing Jin, University of Toronto).

\bibliography{references}

\clearpage
\appendix

\renewcommand{\thefigure}{A\arabic{figure}}
\setcounter{figure}{0}
\renewcommand{\thetable}{A\arabic{table}}
\setcounter{table}{0}
\input{Appendixes/Append_A_InterProfSim}

\renewcommand{\thefigure}{B\arabic{figure}}
\setcounter{figure}{0}
\renewcommand{\thetable}{B\arabic{table}}
\setcounter{table}{0}
\input{Appendixes/Append_B_DecSensAblation}

\renewcommand{\thefigure}{C\arabic{figure}}
\setcounter{figure}{0}
\renewcommand{\thetable}{C\arabic{table}}
\setcounter{table}{0}
\input{Appendixes/Append_C_FullPairStat}

\renewcommand{\thefigure}{D\arabic{figure}}
\setcounter{figure}{0}
\renewcommand{\thetable}{D\arabic{table}}
\setcounter{table}{0}
\input{Appendixes/Append_D_MatchFactConstrast}

\renewcommand{\thefigure}{E\arabic{figure}}
\setcounter{figure}{0}
\renewcommand{\thetable}{E\arabic{table}}
\setcounter{table}{0}
\input{Appendixes/Append_E_MatchJustCaptionContrast}

\renewcommand{\thefigure}{F\arabic{figure}}
\setcounter{figure}{0}
\renewcommand{\thetable}{F\arabic{table}}
\setcounter{table}{0}
\input{Appendixes/Append_F_NoPersona}

\renewcommand{\thefigure}{G\arabic{figure}}
\setcounter{figure}{0}
\renewcommand{\thetable}{G\arabic{table}}
\setcounter{table}{0}
\input{Appendixes/Append_TopicAnalysis}

\renewcommand{\thefigure}{H\arabic{figure}}
\setcounter{figure}{0}
\renewcommand{\thetable}{H\arabic{table}}
\setcounter{table}{0}
\input{Appendixes/Append_H_Prompts}

\end{document}

%% file: Appendixes/Append_A_InterProfSim.tex
\section{Inter-Profile Similarity}\label{app:InterProfSim}

\begin{figure}[H]
    \centering
    \subfloat[Qwen3-VL]{\includegraphics[width=0.47\textwidth]{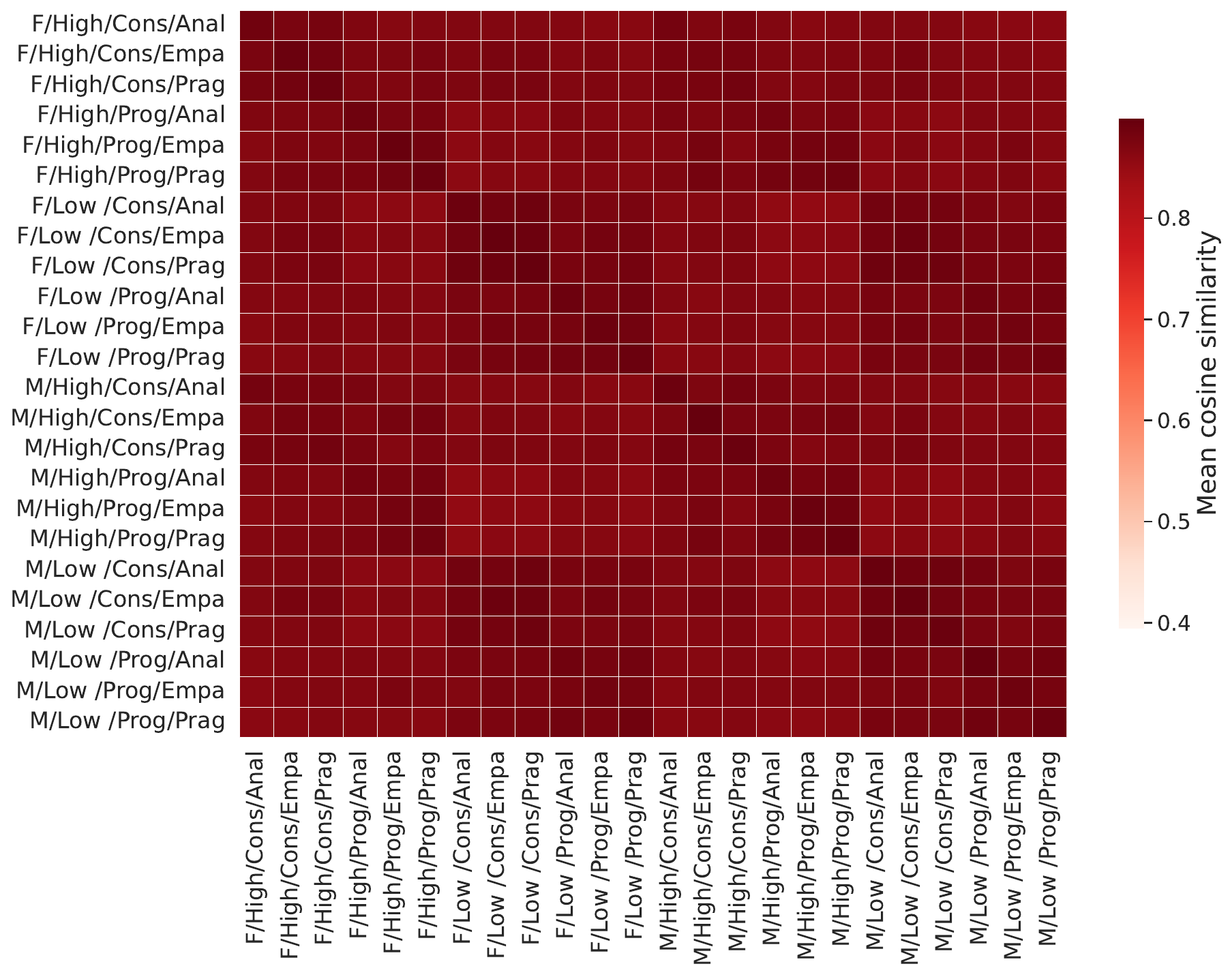}}
    \hfill
    \subfloat[Gemma4]{\includegraphics[width=0.47\textwidth]{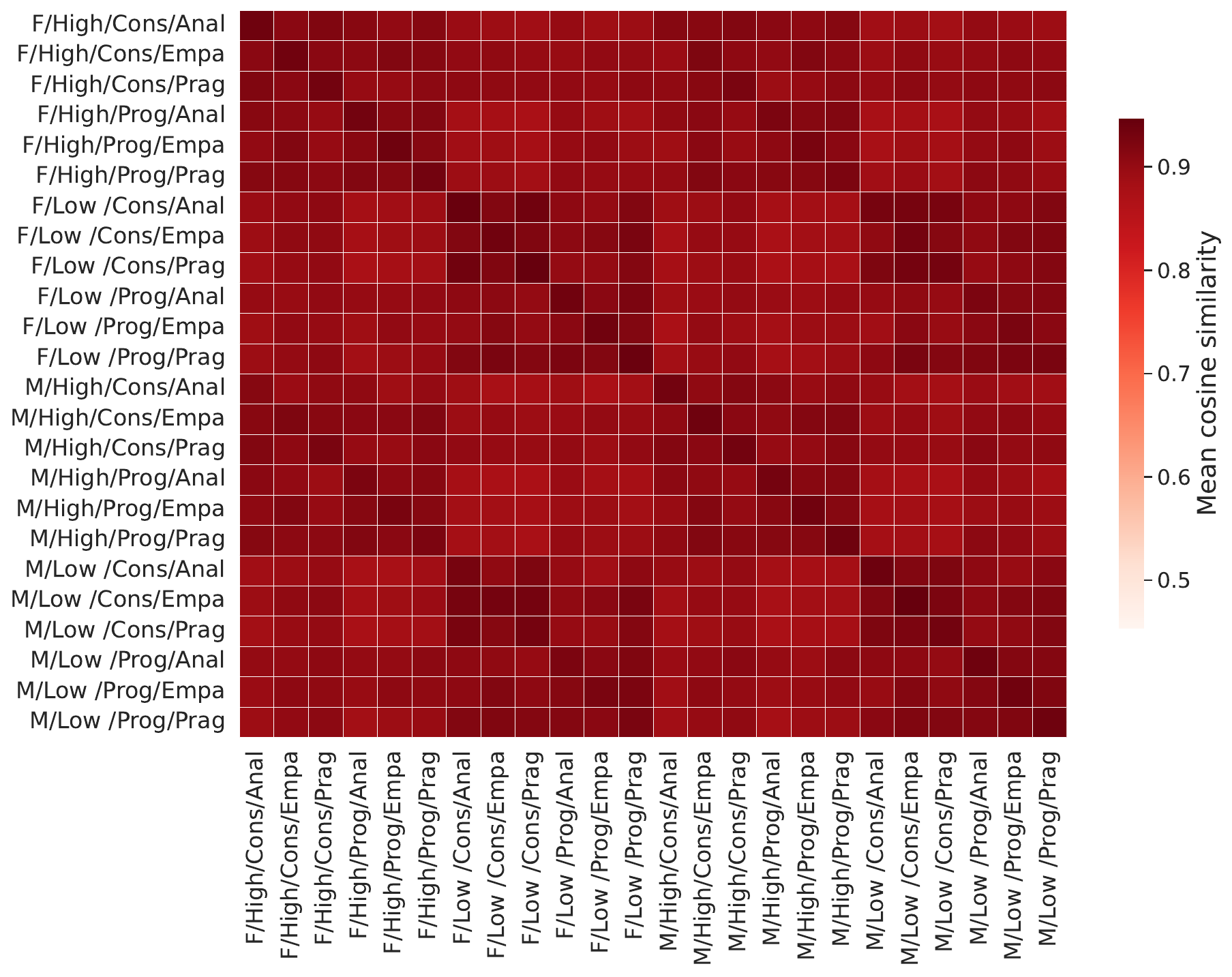}}\vspace{-0.4cm}
    \caption{         Inter-profile \emph{caption} (descriptive grounding) similarity. Notice that, for both models (Qwen3-VL(\ref{fig:ic_caption}a) and Gemma4(\ref{fig:ic_caption}b)), persona conditioning has only a marginal effect on captions (DSI = 2.3\% and 3.7\%, respectively), yielding highly correlated profile-level similarity patterns. The two models agree strongly on this structure (Pearson $r=0.89$; Table~\ref{tab:cross_model_corr}).  
    }
    \label{fig:ic_caption}
\end{figure}

\begin{figure}[H]
    \centering
    \subfloat[Qwen3-VL]{\includegraphics[width=0.47\textwidth]{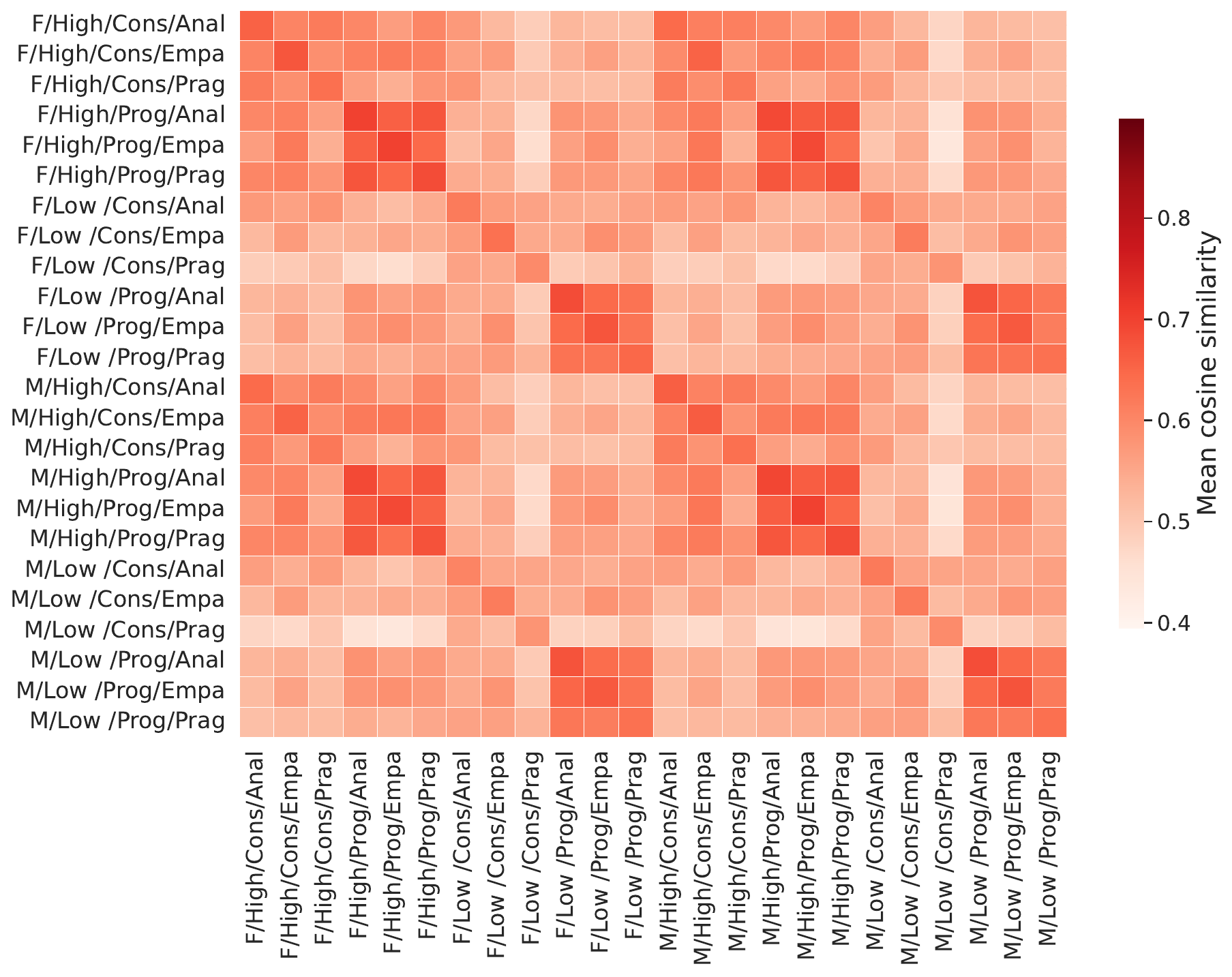}}
    \hfill
    \subfloat[Gemma4]{\includegraphics[width=0.47\textwidth]{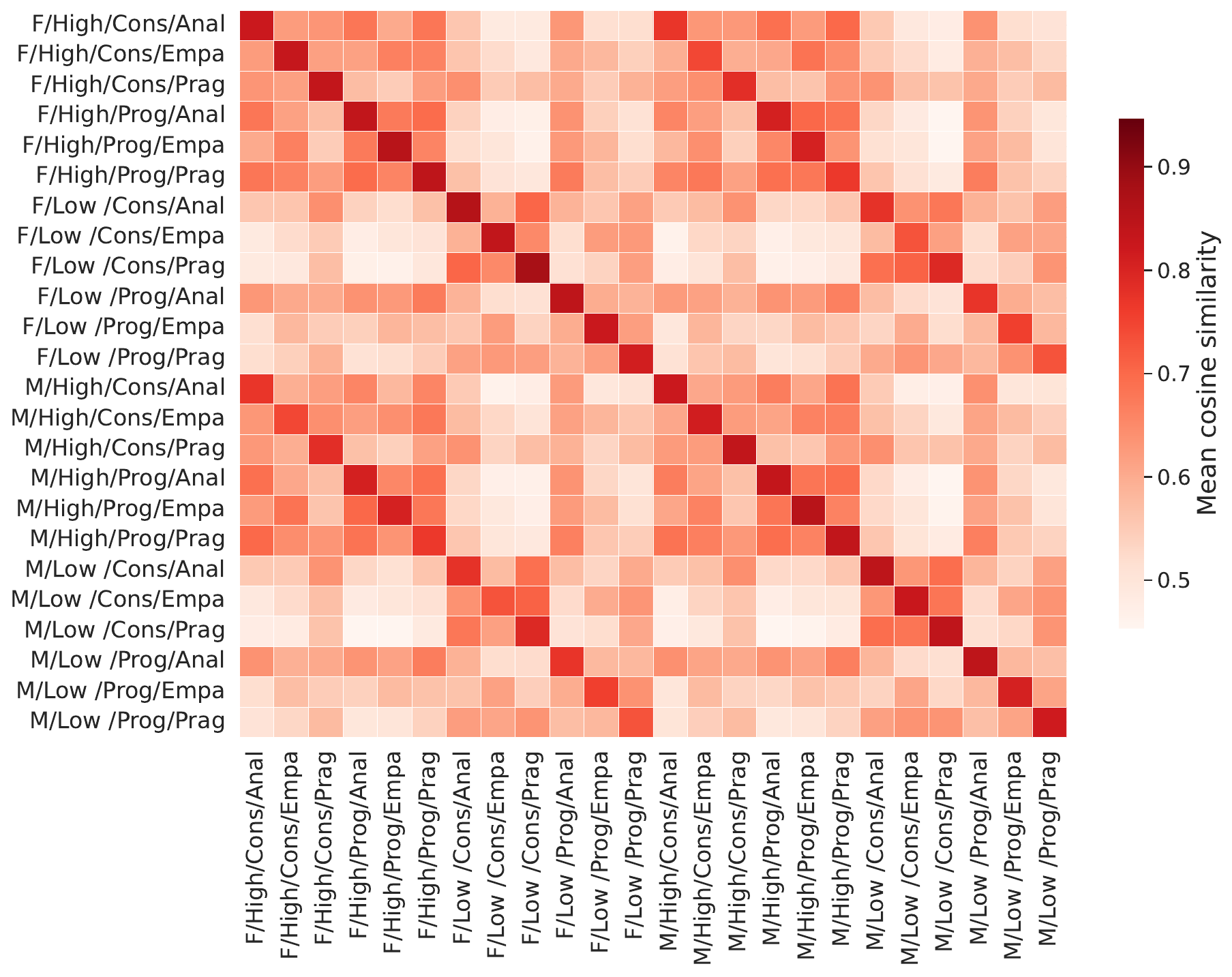}}\vspace{-0.4cm}
    \caption{
        Inter-profile \emph{justification} (interpretive framing) similarity. Both models show a wide range and high-valued diagonal (within-profile) cells: DSI = 14.9\% for Qwen3-VL and 30.0\% for Gemma4) which exhibits a more pronounced structure, yet the two patterns remain strongly correlated (Pearson $r=0.80$).
    }
    \label{fig:ic_just}
\end{figure}

\begin{figure}[htb]
    \centering
    \subfloat[Qwen3-VL]{\includegraphics[width=0.47\textwidth]{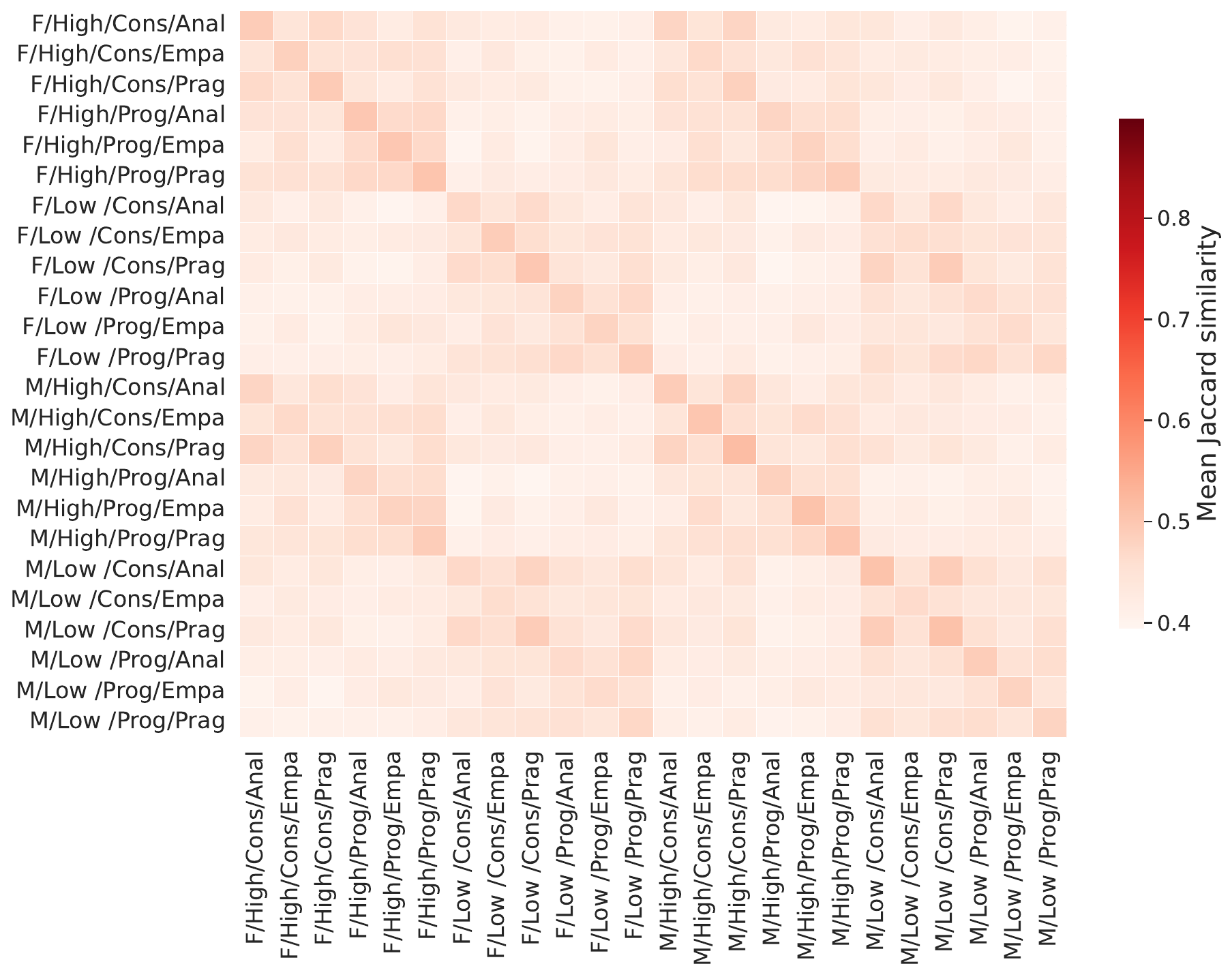}}
    \hfill
    \subfloat[Gemma4]{\includegraphics[width=0.47\textwidth]{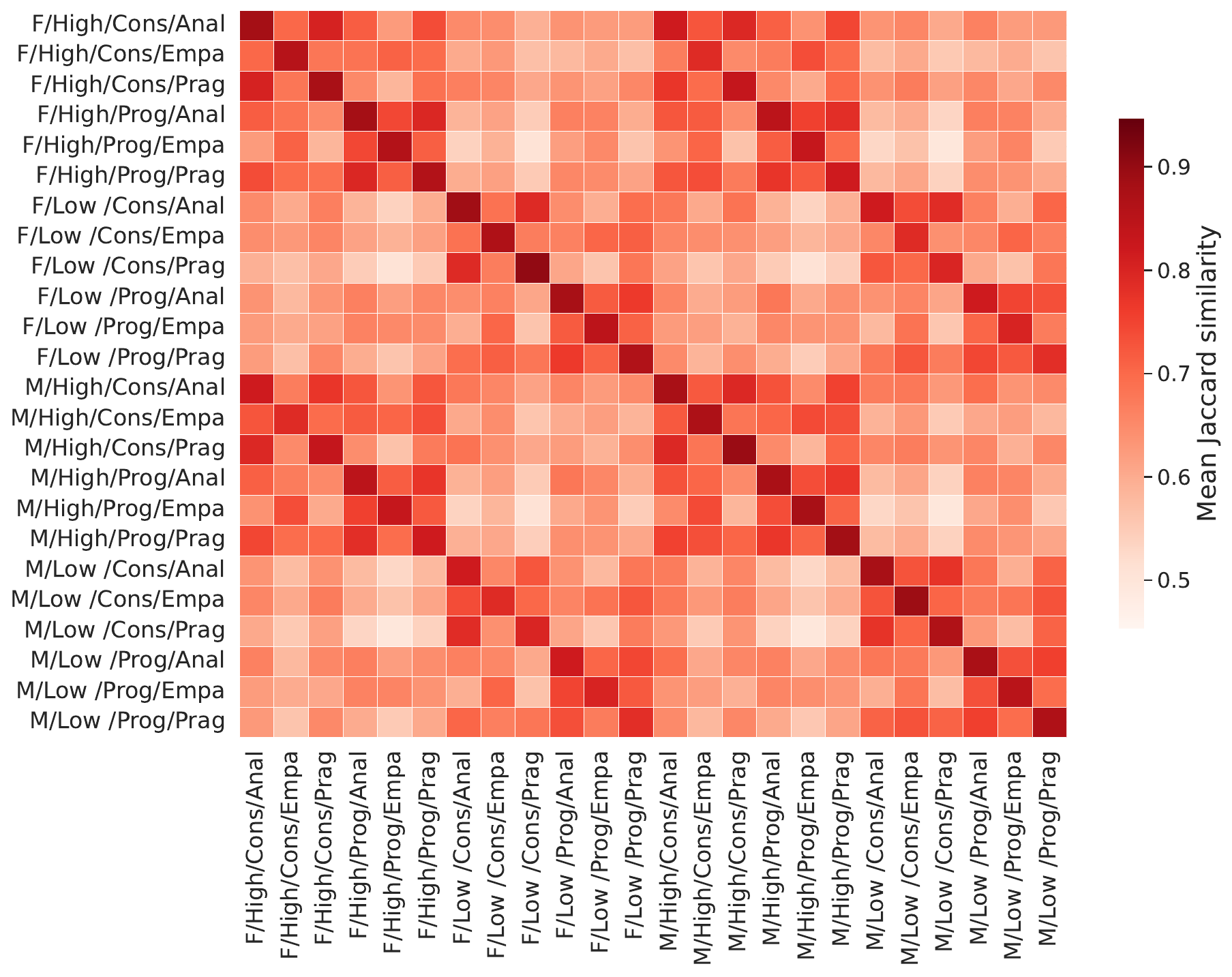}}\vspace{-0.4cm}
    \caption{
        Inter-profile \emph{perception-tag} similarity. DSI = 11.7\% for Qwen3-VL and 25.1\% for Gemma4,  which exhibits higher DSI and similarity values, yet the patterns remain strongly correlated (Pearson $r=0.88$).
    }
    \label{fig:ic_perc}
\end{figure}

This section reports the image-conditioned $24\times24$ similarity matrices for Qwen3-VL and Gemma4, for captions (Figure \ref{fig:ic_caption}) and justifications (Figure \ref{fig:ic_just}), both under cosine scale, and for perceptions (Figure \ref{fig:ic_perc}) under Jaccard scale.

In each matrix, cell $(p_x, p_y)$ is the mean (cosine or Jaccard) similarity between profiles $p_x$ and $p_y$, computed on the same image and averaged over shared images.  Note that the similarity of the perceptions is the \emph{Jaccard} index (intersection over union of the set of controlled-vocabulary tags), so the values lie on a different scale from the cosine heatmaps for the caption and justification cases.

We begin by examining \emph{descriptive grounding}, as captured by captions (Figure~\ref{fig:ic_caption}). For both models, caption similarity spans a narrow range across all cells, including the diagonal ($0.85$--$0.90$ for Qwen3-VL and $0.87$--$0.95$ for Gemma4), with no discernible block structure. For the similarity matrix depicted in Figure~\ref{fig:ic_caption}(a) (Qwen3-VL model), 
$
DSI = \frac{\mu_D-\mu_O}{\mu_D}
=
\frac{0.8928-0.8719}
    {0.8928} 
    =0.0234 \approx 2.3\%
$
and for the similarity matrix depicted in Figure~\ref{fig:ic_caption}(b) (Gemma4 model), $DSI =0.0368 \approx 3.7\%$. This indicates that agents sharing the same profile produce captions that are only marginally more similar than those from different profiles, confirming that descriptive content is largely invariant to persona conditioning in both models. The two caption matrices are strongly correlated (Pearson $r=0.89$; Table~\ref{tab:cross_model_corr}).

Figure  \ref{fig:ic_just} shows that justification similarity spans a wider range (Qwen3-VL: $0.44$--$0.70$; Gemma4: $0.45$--$0.88$) and diagonal cells are consistently among the highest values in their respective rows and columns for both models, confirming that agents within the same profile generate more similar justifications than cross-profile pairs (Figure~\ref{fig:ic_just}). Gemma4 shows an even more pronounced structure (a higher diagonal and a wider off-diagonal spread), but the profiles that resemble each other are shared across models (Pearson $r=0.8$). Thus, persona effects concentrate in interpretive framing rather than descriptive content, consistently across models (highest DSI values for justifications). 
 
Perception tag Jaccard similarity (Figure~\ref{fig:ic_perc}) is reported on a different scale and is not directly comparable to cosine values, as perception tags are controlled-vocabulary sets rather than free text (Section~\ref{secWithPerStability}). Within each model, profiles that diverge most in justification also tend to diverge in perception-tag selection (Qwen3-VL: Pearson $r=0.67$; Gemma4: $r=0.82$), suggesting that differences in interpretive framing are partially reflected in structured semantic selection. The perception-tag matrices are also strongly correlated across models ($r=0.88$; Table~\ref{tab:cross_model_corr}). However, the dimensional analysis shows that the magnitudes of these attribute-associated differences vary substantially across output types and models (Section~\ref{sec:concordance}).

%% file: Appendixes/Append_B_DecSensAblation.tex
\section{Decoding-Sensitivity Ablation at
\texorpdfstring{$T=0$}{T=0}}
\label{app:temperature0}

To test whether the strongest persona-associated pattern identified in the main analysis depends on stochastic decoding, we conduct a controlled ablation at $T=0$. We focus on economic status, which produces the largest justification difference in both models in the main analysis. We hold the remaining attributes fixed as Female, Conservative, and Analytical and vary only the economic-status label (Low vs.\ High). For each model, we generate outputs under both conditions for all $|I|=50$ images. 
In this ablation, both models run through Hugging Face \texttt{transformers} with bf16 weights (Qwen3-VL as \texttt{Qwen/Qwen3-VL-8B-Instruct}), so for Qwen3-VL the ablation also differs from the main corpus in serving stack and precision.

Under the evaluated deterministic configuration, repeated generations within each economic-status condition produced byte-identical outputs
because the instance identifier was not included in the prompt. Within-condition similarity was therefore 1.000 and is not informative
for this analysis. Instead, we compute, for each image, the cross-condition similarity between the High- and Low-income outputs. We use cosine
similarity for captions and justifications and Jaccard similarity for perception tags, following Section~\ref{secWithPerStability}.

\begin{figure}[t]
    \centering
    \includegraphics[width=0.85\columnwidth]
    {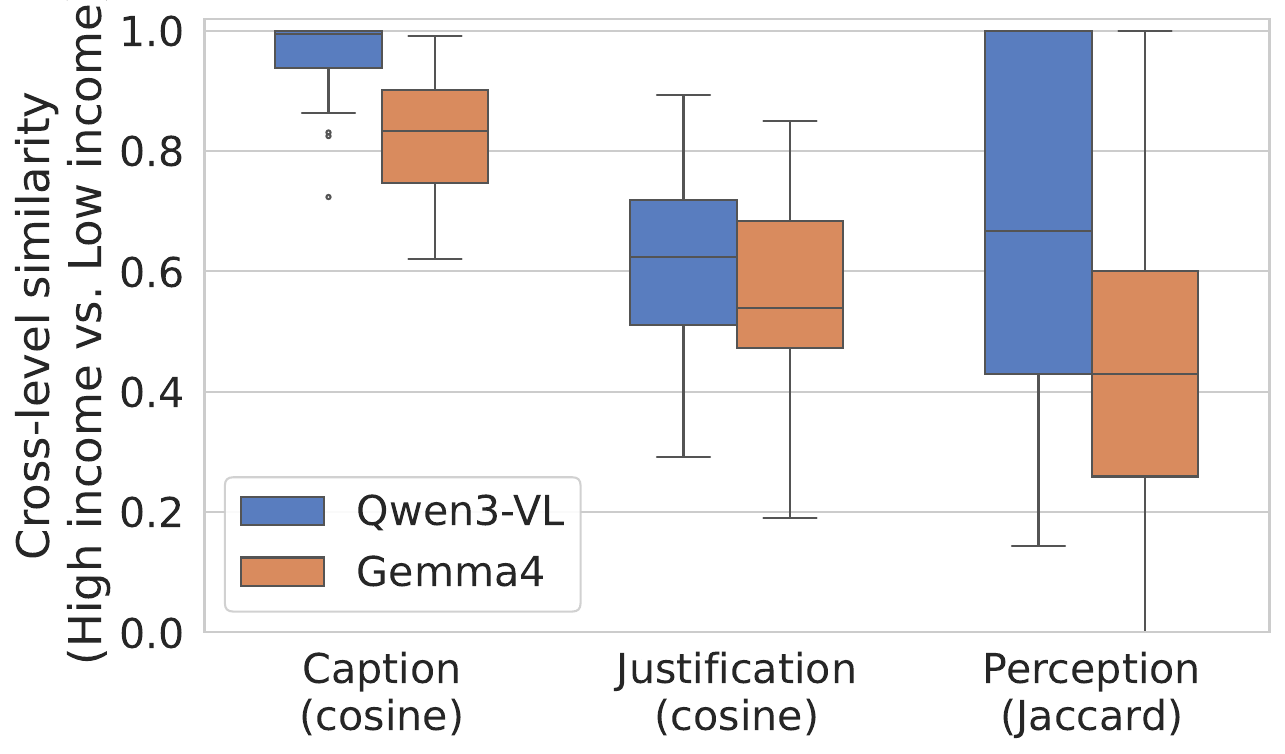}
    \vspace{-0.4cm}
    \caption{
    $T=0$ ablation: distributions of per-image cross-condition similarity between outputs generated with the High- and Low-income persona labels for Qwen3-VL and Gemma4, with all other persona attributes held fixed. Caption and justification values use cosine similarity, whereas perception tags use Jaccard similarity.
    }
    \label{fig:t0_ablation}
\end{figure}

Changing the economic-status label leaves captions unchanged in 24 of 50 images (48\%) for Qwen3-VL, whereas no Gemma4 caption pairs are
byte-identical. Justifications differ for every image in both models (see Figure \ref{fig:t0_ablation}). Nevertheless, caption similarity remains high across economic status conditions (Qwen3-VL: 0.960; Gemma4: 0.823), whereas justification similarity is substantially lower (0.612 and 0.554, respectively). The corresponding mean caption-minus-justification similarity differences are 0.348 for Qwen3-VL and 0.269 for Gemma4. Paired Wilcoxon signed-rank tests over the 50 images confirm that cross-condition similarity is lower for justifications than for captions in both models (Qwen3-VL: $p=1.8\times10^{-15}$; Gemma4: $p=2.2\times10^{-12}$).

Perception-tag Jaccard similarity is 0.673 for Qwen3-VL and 0.423 for Gemma4. Because perception tags use Jaccard similarity rather than the cosine similarity used for captions and justifications, these values are interpreted separately.

Thus, the economic-status caption--justification contrast persists under deterministic decoding in both models. Changing the economic-status label affects interpretive framing more strongly than descriptive grounding, even when stochastic sampling is removed, supporting the robustness of the main economic-status result to decoding temperature.

%% file: Appendixes/Append_C_FullPairStat.tex
\section{Full Paired Attribute-Level Statistics}
\label{app:stats}

Table~\ref{tab:within_cross_qwenvl_gemma4} reports the full paired image-level results for both models. Panel A reports the mean within-minus-cross difference across the images, $\Delta_d$, for each persona dimension and output type. Panel B reports the corresponding justification-minus-caption contrast. The terms in Panel B are directly comparable because captions and justifications are represented using the same embedding model and evaluated using cosine similarity.

For each comparison, $n_{+}$ denotes the number of images with a positive paired difference. Let $W_{+}$ and $W_{-}$ denote the sums of the positive and negative signed ranks, respectively. We report $W=\min(W_{+},W_{-})$ and the matched-pairs rank-biserial correlation $r_{\mathrm{rb}}=(W_{+}-W_{-})/(W_{+}+W_{-})$. Mean differences and their $95\%$ bias-corrected and accelerated (BCa) bootstrap confidence intervals summarize effect magnitude, while $p_{\mathrm{BH}}$ reports the two-sided Wilcoxon signed-rank result after Benjamini--Hochberg correction.

\begin{table*}[tbh]
\centering
\scriptsize
\setlength{\tabcolsep}{4pt}
\renewcommand{\arraystretch}{1.15}
\resizebox{\textwidth}{!}{%
\begin{tabular}{@{}ll rrrrr @{\hspace{14pt}} rrrrr@{}}
\toprule
& & \multicolumn{5}{c}{\textbf{Qwen3-VL}} & \multicolumn{5}{c}{\textbf{Gemma4}} \\
\cmidrule(lr){3-7}\cmidrule(lr){8-12}
Type & Dimension &  Estimate [95\% BCa CI] & $n_{+}$ & $W$ & $r_{\mathrm{rb}}$ & $p_{\mathrm{BH}}$ &  Estimate [95\% BCa CI] & $n_{+}$ & $W$ & $r_{\mathrm{rb}}$ & $p_{\mathrm{BH}}$ \\
\midrule
\multicolumn{12}{@{}l}{\textbf{Panel A.}\enspace\textit{Within-group minus cross-group similarity}} \\
\addlinespace[2pt]
\multirow{4}{*}{Caption} & Economic status & $+0.0138$\;{\tiny\textcolor{black!55}{$[+0.0102,\,+0.0224]$}} & 50 & 0 & $1.000$ & $2.13\times10^{-15}$ & $+0.0211$\;{\tiny\textcolor{black!55}{$[+0.0159,\,+0.0282]$}} & 50 & 0 & $1.000$ & $2.37\times10^{-15}$ \\
 & Political orientation & $+0.0072$\;{\tiny\textcolor{black!55}{$[+0.0035,\,+0.0217]$}} & 50 & 0 & $1.000$ & $2.13\times10^{-15}$ & $+0.0089$\;{\tiny\textcolor{black!55}{$[+0.0059,\,+0.0144]$}} & 49 & 1 & $0.998$ & $4.26\times10^{-15}$ \\
 & Personality & $+0.0032$\;{\tiny\textcolor{black!55}{$[+0.0025,\,+0.0043]$}} & 50 & 0 & $1.000$ & $2.13\times10^{-15}$ & $+0.0070$\;{\tiny\textcolor{black!55}{$[+0.0056,\,+0.0086]$}} & 50 & 0 & $1.000$ & $2.37\times10^{-15}$ \\
 & Gender & $+0.0017$\;{\tiny\textcolor{black!55}{$[+0.0012,\,+0.0028]$}} & 50 & 0 & $1.000$ & $2.13\times10^{-15}$ & $+0.0024$\;{\tiny\textcolor{black!55}{$[+0.0018,\,+0.0034]$}} & 46 & 10 & $0.984$ & $7.64\times10^{-14}$ \\
\addlinespace
\multirow{4}{*}{Justification} & Economic status & $+0.0617$\;{\tiny\textcolor{black!55}{$[+0.0538,\,+0.0710]$}} & 50 & 0 & $1.000$ & $2.13\times10^{-15}$ & $+0.1015$\;{\tiny\textcolor{black!55}{$[+0.0922,\,+0.1127]$}} & 50 & 0 & $1.000$ & $2.37\times10^{-15}$ \\
 & Political orientation & $+0.0444$\;{\tiny\textcolor{black!55}{$[+0.0390,\,+0.0520]$}} & 50 & 0 & $1.000$ & $2.13\times10^{-15}$ & $+0.0561$\;{\tiny\textcolor{black!55}{$[+0.0500,\,+0.0639]$}} & 50 & 0 & $1.000$ & $2.37\times10^{-15}$ \\
 & Personality & $+0.0225$\;{\tiny\textcolor{black!55}{$[+0.0200,\,+0.0257]$}} & 50 & 0 & $1.000$ & $2.13\times10^{-15}$ & $+0.0632$\;{\tiny\textcolor{black!55}{$[+0.0583,\,+0.0699]$}} & 50 & 0 & $1.000$ & $2.37\times10^{-15}$ \\
 & Gender & $+0.0021$\;{\tiny\textcolor{black!55}{$[+0.0018,\,+0.0026]$}} & 50 & 0 & $1.000$ & $2.13\times10^{-15}$ & $+0.0073$\;{\tiny\textcolor{black!55}{$[+0.0064,\,+0.0084]$}} & 50 & 0 & $1.000$ & $2.37\times10^{-15}$ \\
\addlinespace
\multirow{4}{*}{Perception} & Economic status & $+0.0358$\;{\tiny\textcolor{black!55}{$[+0.0273,\,+0.0479]$}} & 49 & 1 & $0.998$ & $3.88\times10^{-15}$ & $+0.1068$\;{\tiny\textcolor{black!55}{$[+0.0853,\,+0.1393]$}} & 50 & 0 & $1.000$ & $2.37\times10^{-15}$ \\
 & Political orientation & $+0.0171$\;{\tiny\textcolor{black!55}{$[+0.0137,\,+0.0214]$}} & 50 & 0 & $1.000$ & $2.13\times10^{-15}$ & $+0.0650$\;{\tiny\textcolor{black!55}{$[+0.0481,\,+0.0920]$}} & 50 & 0 & $1.000$ & $2.37\times10^{-15}$ \\
 & Personality & $+0.0122$\;{\tiny\textcolor{black!55}{$[+0.0097,\,+0.0156]$}} & 50 & 0 & $1.000$ & $2.13\times10^{-15}$ & $+0.0469$\;{\tiny\textcolor{black!55}{$[+0.0379,\,+0.0572]$}} & 50 & 0 & $1.000$ & $2.37\times10^{-15}$ \\
 & Gender & $+0.0030$\;{\tiny\textcolor{black!55}{$[+0.0023,\,+0.0040]$}} & 48 & 5 & $0.992$ & $1.78\times10^{-14}$ & $+0.0099$\;{\tiny\textcolor{black!55}{$[+0.0074,\,+0.0133]$}} & 49 & 6 & $0.991$ & $2.71\times10^{-14}$ \\
\midrule
\multicolumn{12}{@{}l}{\textbf{Panel B.}\enspace\textit{Justification minus caption within--cross difference}} \\
\addlinespace[2pt]
\multirow{4}{*}{Justification $-$ caption} & Economic status & $+0.0478$\;{\tiny\textcolor{black!55}{$[+0.0389,\,+0.0575]$}} & 49 & 29 & $0.955$ & $6.18\times10^{-12}$ & $+0.0804$\;{\tiny\textcolor{black!55}{$[+0.0707,\,+0.0913]$}} & 50 & 0 & $1.000$ & $2.37\times10^{-15}$ \\
 & Political orientation & $+0.0372$\;{\tiny\textcolor{black!55}{$[+0.0307,\,+0.0438]$}} & 49 & 40 & $0.937$ & $2.06\times10^{-11}$ & $+0.0472$\;{\tiny\textcolor{black!55}{$[+0.0416,\,+0.0537]$}} & 50 & 0 & $1.000$ & $2.37\times10^{-15}$ \\
 & Personality & $+0.0193$\;{\tiny\textcolor{black!55}{$[+0.0167,\,+0.0225]$}} & 50 & 0 & $1.000$ & $7.11\times10^{-15}$ & $+0.0562$\;{\tiny\textcolor{black!55}{$[+0.0512,\,+0.0632]$}} & 50 & 0 & $1.000$ & $2.37\times10^{-15}$ \\
 & Gender & $+0.0004$\textsuperscript{\dag}\;{\tiny\textcolor{black!55}{$[-0.0003,\,+0.0009]$}} & 36 & 328 & $0.485$ & $0.00236$ & $+0.0049$\;{\tiny\textcolor{black!55}{$[+0.0036,\,+0.0062]$}} & 45 & 90 & $0.859$ & $3.71\times10^{-9}$ \\
\bottomrule
\end{tabular}%
}\vspace{-0.3cm}
\caption{
Paired image-level analysis for both MLLMs.  Panel A reports the within-minus-cross difference $\Delta_d$, and Panel B reports the justification-minus-caption contrast $\Delta_d^{\mathrm{just}}-\Delta_d^{\mathrm{cap}}$. $n_{+}$ is the number of images with a positive paired difference, $W$ is the smaller signed-rank sum, and $r_{\mathrm{rb}}$ is the matched-pairs rank-biserial correlation. The displayed $p$-values are from two-sided Wilcoxon signed-rank tests with Benjamini--Hochberg correction applied separately within each model and panel: 12 comparisons in Panel A and four in Panel B. Confidence intervals are $95\%$ BCa bootstrap intervals for the mean image-level difference. Because the confidence intervals and signed-rank tests target different summary quantities, they need not always yield identical conclusions. $\dagger$ marks a confidence interval that includes zero. 
}
\label{tab:within_cross_qwenvl_gemma4}
\end{table*}

All signed-rank results in Table~\ref{tab:within_cross_qwenvl_gemma4} remain significant after Benjamini--Hochberg correction within their respective testing families. Because statistical significance is nearly uniform across comparisons, substantive interpretation focuses on the mean differences and their BCa confidence intervals. In particular, the Qwen3-VL gender contrast in Panel B has a small mean of $+0.0004$ and a confidence interval that includes zero, despite a significant signed-rank result. This difference arises because the BCa interval summarizes the arithmetic mean, whereas the signed-rank test evaluates the ranked location of the paired differences.

Exact two-sided sign tests provide an additional sensitivity analysis and also remain significant after correction. The largest adjusted sign-test $p$-values are $4.5\times10^{-10}$ in Panel A and $0.00260$ in Panel B.

A complete case sensitivity analysis yields the same substantive conclusions. For Qwen3-VL, restricting the analysis to the 33 images with all $1{,}200$ persona-conditioned generations leaves the mean gender justification-minus-caption contrast unchanged at $+0.0004$, although its adjusted signed-rank result becomes $p_{\mathrm{BH}}=0.0801$. The other three Qwen3-VL contrasts remain substantively unchanged. For Gemma4, all four contrasts remain substantively unchanged when the analysis is restricted to its 49 complete images.

\textbf{Reading example.}
For Qwen3-VL economic status, $\Delta_d=+0.0617$ indicates that justifications generated under profiles sharing the same economic status label are, on average, $0.0617$ cosine points more similar than justifications generated under profiles with different economic status labels. The $95\%$ BCa interval $[+0.0538,+0.0710]$ excludes zero, and all 50 image-level differences are positive.

%% file: Appendixes/Append_D_MatchFactConstrast.tex
\section{Matched-Factorial Contrast}
\label{app:matched_factorial}

The analysis in Appendix~\ref{app:stats} compares pairs that share or differ on a target persona dimension while averaging over the remaining three dimensions. Consequently, profiles in a cross-group pair may also differ on one or more additional attributes. As a complementary controlled analysis, we repeat the comparison using matched profile pairs that agree on all other attributes and differ only on the target dimension.

The four persona dimensions define 24 complete profiles. For a target dimension $d$, let $\mathcal{M}_d$ denote the set of profile pairs $(p,q)$ that agree on every other attribute and differ only on $d$. This produces 12 matched pairs for each binary dimension. For personality, the eight combinations of the remaining binary attributes each contribute three pairwise comparisons among Pragmatic, Empathetic, and Analytical, producing 24 matched pairs.

For image $i$ and matched pair $(p,q)\in\mathcal{M}_d$, we define
\[
\delta_{i,pq,d}
=
\frac{1}{2}
\left(
s^{\mathrm{\odot}}_{i,p}
+
s^{\mathrm{\odot}}_{i,q}
\right)
-
s^{\mathrm{\otimes}}_{i,pq},
\]
where $s^{\mathrm{\odot}}_{i,p}$ is the mean pairwise similarity among outputs generated under profile $p$ for image $i$, and $s^{\mathrm{\otimes}}_{i,pq}$ is the mean similarity between outputs generated under profiles $p$ and $q$ for that image. We then average over the matched profile pairs:
\[
\Delta^{\mathrm{match}}_{i,d}
=
\frac{1}{|\mathcal{M}_d|}
\sum_{(p,q)\in\mathcal{M}_d}
\delta_{i,pq,d}.
\]

This procedure yields one matched-factorial contrast for each of the 50 images. We apply the same image-level inferential procedure used in Appendix~\ref{app:stats}: a two-sided Wilcoxon signed-rank test, a 95\% BCa bootstrap confidence interval obtained by resampling images, an exact sign-test sensitivity analysis, and Benjamini--Hochberg correction across the 12 comparisons within each model. Similarities are computed using all available valid generations. Every image--profile cell contains at least two valid outputs, so all within-profile and between-profile similarities are defined.

Table~\ref{tab:matched_factorial} reports the matched-factorial estimates. The economic-status dimension produces the largest matched contrast and gender the smallest across both models and all three output types. The relative ordering of the four dimensions is preserved in five of the six model--output type combinations. For Gemma4 justifications, political orientation slightly exceeds personality ($+0.2037$ versus $+0.1976$), reversing their second- and third-ranked positions in the marginal analysis.

\begin{table*}[tbh]
\centering
\scriptsize
\setlength{\tabcolsep}{6pt}
\renewcommand{\arraystretch}{1.15}
\begin{tabular}{@{}llrcc@{}}
\toprule
& & &
\multicolumn{2}{c}{Mean $\Delta^{\mathrm{match}}$
[95\% BCa CI]} \\
\cmidrule(lr){4-5}
Type & Dimension & Pairs & Qwen3-VL & Gemma4 \\
\midrule

\multirow{4}{*}{Caption}
& Economic status
& 12
& $+0.0229$\;{\tiny\textcolor{black!55}{$[+0.0180,\,+0.0319]$}}
& $+0.0380$\;{\tiny\textcolor{black!55}{$[+0.0305,\,+0.0473]$}}
\\

& Political orientation
& 12
& $+0.0156$\;{\tiny\textcolor{black!55}{$[+0.0110,\,+0.0306]$}}
& $+0.0239$\;{\tiny\textcolor{black!55}{$[+0.0179,\,+0.0324]$}}
\\

& Personality
& 24
& $+0.0098$\;{\tiny\textcolor{black!55}{$[+0.0079,\,+0.0126]$}}
& $+0.0199$\;{\tiny\textcolor{black!55}{$[+0.0165,\,+0.0239]$}}
\\

& Gender
& 12
& $+0.0074$\;{\tiny\textcolor{black!55}{$[+0.0058,\,+0.0099]$}}
& $+0.0112$\;{\tiny\textcolor{black!55}{$[+0.0088,\,+0.0147]$}}
\\

\addlinespace
\multirow{4}{*}{Justification}
& Economic status
& 12
& $+0.0976$\;{\tiny\textcolor{black!55}{$[+0.0876,\,+0.1084]$}}
& $+0.2684$\;{\tiny\textcolor{black!55}{$[+0.2501,\,+0.2861]$}}
\\

& Political orientation
& 12
& $+0.0805$\;{\tiny\textcolor{black!55}{$[+0.0732,\,+0.0903]$}}
& $+0.2037$\;{\tiny\textcolor{black!55}{$[+0.1877,\,+0.2202]$}}
\\

& Personality
& 24
& $+0.0456$\;{\tiny\textcolor{black!55}{$[+0.0414,\,+0.0507]$}}
& $+0.1976$\;{\tiny\textcolor{black!55}{$[+0.1858,\,+0.2098]$}}
\\

& Gender
& 12
& $+0.0106$\;{\tiny\textcolor{black!55}{$[+0.0095,\,+0.0117]$}}
& $+0.0662$\;{\tiny\textcolor{black!55}{$[+0.0609,\,+0.0726]$}}
\\

\addlinespace
\multirow{4}{*}{Perception}
& Economic status
& 12
& $+0.0619$\;{\tiny\textcolor{black!55}{$[+0.0518,\,+0.0758]$}}
& $+0.2349$\;{\tiny\textcolor{black!55}{$[+0.1950,\,+0.2778]$}}
\\

& Political orientation
& 12
& $+0.0413$\;{\tiny\textcolor{black!55}{$[+0.0363,\,+0.0479]$}}
& $+0.1749$\;{\tiny\textcolor{black!55}{$[+0.1425,\,+0.2134]$}}
\\

& Personality
& 24
& $+0.0325$\;{\tiny\textcolor{black!55}{$[+0.0282,\,+0.0375]$}}
& $+0.1411$\;{\tiny\textcolor{black!55}{$[+0.1169,\,+0.1659]$}}
\\

& Gender
& 12
& $+0.0171$\;{\tiny\textcolor{black!55}{$[+0.0147,\,+0.0197]$}}
& $+0.0619$\;{\tiny\textcolor{black!55}{$[+0.0496,\,+0.0773]$}}
\\

\bottomrule
\end{tabular}

\caption{
Matched-factorial contrasts for both MLLMs. Each row averages over profile pairs that agree on the other three persona attributes and differ only on the stated dimension. For each matched pair, the contrast compares the average within-profile similarity of the two profiles with their between-profile similarity. \emph{Pairs} denotes the number of matched profile pairs. All 50 image-level contrasts are positive in every comparison, yielding $n_{+}=50$, $W=0$, $r_{\mathrm{rb}}=1.000$, and $p_{\mathrm{BH}}=1.78\times10^{-15}$ throughout; these repeated statistics are omitted from the table. Exact sign tests also remain significant after Benjamini--Hochberg correction. Caption and justification results use cosine similarity, whereas perception-tag results use Jaccard similarity and are interpreted separately.
}
\label{tab:matched_factorial}
\end{table*}

The marginal and matched analyses provide complementary views of persona-associated variation. Appendix~\ref{app:stats} summarizes within-group and cross-group similarity across the full set of profiles, averaging over the remaining persona dimensions. The matched-factorial analysis instead isolates each target dimension by comparing profiles that are identical on the other three attributes. Because the two analyses use different profile-pair sets and different within-group reference quantities, their numerical magnitudes should not be compared directly.

The matched estimator also incorporates within-profile reproducibility through $s^{\mathrm{\odot}}_{i,p}$, which compares outputs generated under the same complete persona prompt. Because all image-level matched contrasts are positive, their magnitudes and the relative ordering of dimensions provide the most informative comparisons. The economic-status dimension remains the largest, and gender remains the smallest across models and output types.

The marginal analysis remains the primary presentation because it summarizes attribute-associated similarity across the full corpus and corresponds directly to the distributions in Figure~\ref{fig:within_cross}. The matched-factorial analysis provides complementary controlled evidence that the main dimensional pattern persists when the other three persona attributes are held fixed.

%% file: Appendixes/Append_E_MatchJustCaptionContrast.tex
\section{Matched Justification--Caption Contrast}
\label{app:matched_factorial_contrast}

Table~\ref{tab:matched_factorial} reports the matched caption and justification differences separately. Because both output types are represented using the same embedding model and evaluated with cosine similarity, their matched differences are directly comparable. For each image $i$ and persona dimension $d$, we define
\[
\Gamma_{i,d}
=
\Delta^{\mathrm{just},\mathrm{match}}_{i,d}
-
\Delta^{\mathrm{cap},\mathrm{match}}_{i,d}.
\]
Positive values indicate that the matched attribute-associated difference is larger for justifications than for captions.

We analyze the 50 image-level contrasts using a two-sided Wilcoxon signed-rank test and report a 95\% bias-corrected and accelerated (BCa) bootstrap confidence interval for the mean contrast, obtained by resampling images. Benjamini--Hochberg correction is applied across the four persona dimensions within each model. An exact two-sided sign test provides an additional sensitivity analysis. Perception tags are excluded because their Jaccard similarities are not commensurate with the cosine similarities used for captions and justifications.

\begin{table*}[tbh]
\centering
\scriptsize
\setlength{\tabcolsep}{4pt}
\renewcommand{\arraystretch}{1.15}
\resizebox{\textwidth}{!}{%
\begin{tabular}{@{}l rrrrr @{\hspace{10pt}} rrrrr@{}}
\toprule
& \multicolumn{5}{c}{\textbf{Qwen3-VL}} & \multicolumn{5}{c}{\textbf{Gemma4}} \\
\cmidrule(lr){2-6}\cmidrule(lr){7-11}
Dimension & Mean $\Gamma$ [95\% BCa CI] & $n_{+}$ & $W$ & $r_{\mathrm{rb}}$ & $p_{\mathrm{BH}}$ & Mean $\Gamma$ [95\% BCa CI] & $n_{+}$ & $W$ & $r_{\mathrm{rb}}$ & $p_{\mathrm{BH}}$ \\
\midrule
Economic status & $+0.0747$\;{\tiny\textcolor{black!55}{$[+0.0638,\,+0.0862]$}} & 49 & 5 & $0.992$ & $2.37\times10^{-14}$ & $+0.2303$\;{\tiny\textcolor{black!55}{$[+0.2119,\,+0.2489]$}} & 50 & 0 & $1.000$ & $1.78\times10^{-15}$ \\
Political orientation & $+0.0649$\;{\tiny\textcolor{black!55}{$[+0.0567,\,+0.0739]$}} & 49 & 4 & $0.994$ & $2.37\times10^{-14}$ & $+0.1799$\;{\tiny\textcolor{black!55}{$[+0.1638,\,+0.1962]$}} & 50 & 0 & $1.000$ & $1.78\times10^{-15}$ \\
Personality & $+0.0358$\;{\tiny\textcolor{black!55}{$[+0.0315,\,+0.0413]$}} & 50 & 0 & $1.000$ & $7.11\times10^{-15}$ & $+0.1777$\;{\tiny\textcolor{black!55}{$[+0.1659,\,+0.1900]$}} & 50 & 0 & $1.000$ & $1.78\times10^{-15}$ \\
Gender & $+0.0032$\;{\tiny\textcolor{black!55}{$[+0.0012,\,+0.0048]$}} & 38 & 270 & $0.576$ & $2.52\times10^{-4}$ & $+0.0550$\;{\tiny\textcolor{black!55}{$[+0.0491,\,+0.0617]$}} & 50 & 0 & $1.000$ & $1.78\times10^{-15}$ \\
\bottomrule
\end{tabular}%
}
\caption{
Matched-factorial justification-minus-caption contrasts,
$\Gamma_{i,d}
=
\Delta^{\mathrm{just},\mathrm{match}}_{i,d}
-
\Delta^{\mathrm{cap},\mathrm{match}}_{i,d}$. Positive values indicate that the matched attribute-associated difference is larger for justifications than for captions. Each row uses 50 image-level contrasts. Confidence intervals are 95\% bias-corrected and accelerated bootstrap intervals for the mean,
obtained by resampling images. The displayed $p_{\mathrm{BH}}$ values are from two-sided Wilcoxon signed-rank tests with Benjamini--Hochberg correction across the four dimensions within each model. $n_{+}$ is the number of positive image-level contrasts, $W$ is the smaller signed-rank sum, and $r_{\mathrm{rb}}$ is the matched-pairs rank-biserial correlation. Exact two-sided sign tests also remain significant after correction.
}
\label{tab:matched_factorial_contrast}
\end{table*}

Table~\ref{tab:matched_factorial_contrast} reports the resulting contrasts. All mean values are positive, ranging from $+0.0032$ to $+0.0747$ for Qwen3-VL and from $+0.0550$ to $+0.2303$ for Gemma4. Economic status produces the largest justification-minus-caption contrast and gender the smallest in both models. The confidence intervals exclude zero for all eight model--dimension comparisons, and all Wilcoxon results remain significant after correction.

The Qwen3-VL gender result provides a useful sensitivity case. In the marginal analysis, the mean justification-minus-caption contrast is $+0.0004$, with a confidence interval that includes zero (Table~\ref{tab:within_cross_qwenvl_gemma4}). In the matched design, the mean contrast remains small but is positive at $+0.0032$, with a 95\% BCa interval of $[+0.0012,+0.0048]$. Thus, the broader pattern of greater attribute-associated variation in justifications than in captions persists when the remaining persona attributes are held fixed. Across both models, the matched analysis also preserves the
main ordering, with economic status producing the largest contrast and gender the smallest.

%% file: Appendixes/Append_F_NoPersona.tex
\begin{figure*}[!tb]
    \centering
    \subfloat[Captions, NPT]{%
    \includegraphics[width=0.235\textwidth]{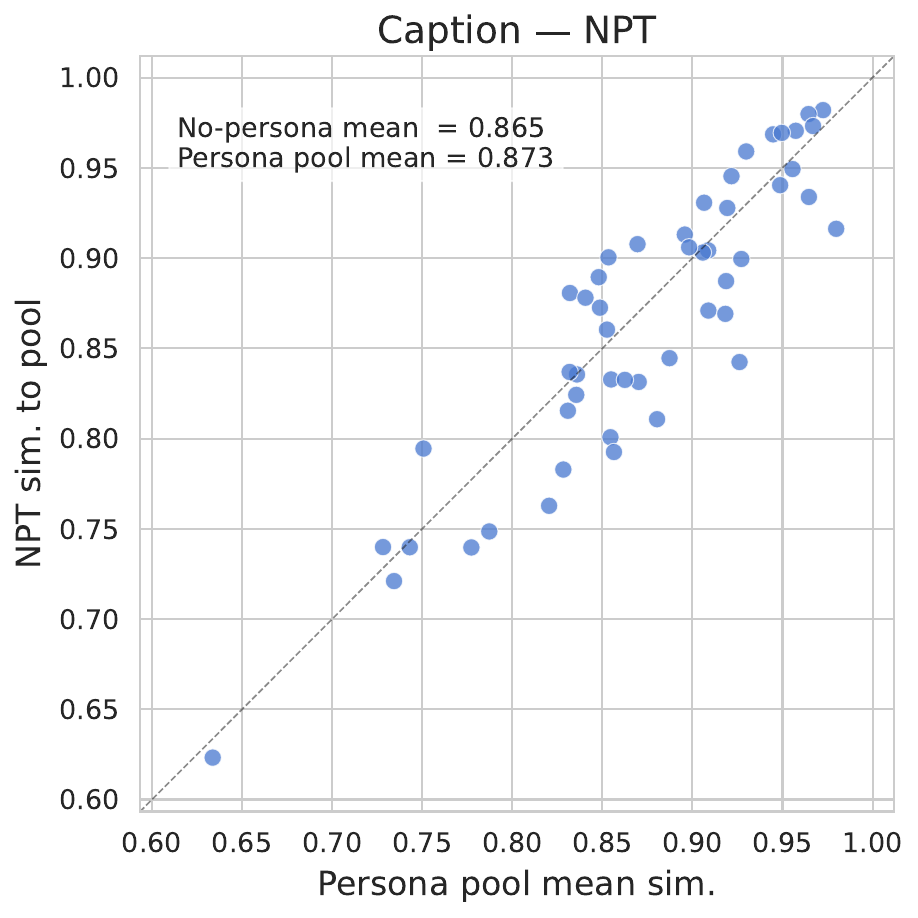}%
    \hspace{2pt}%
    \includegraphics[width=0.235\textwidth]{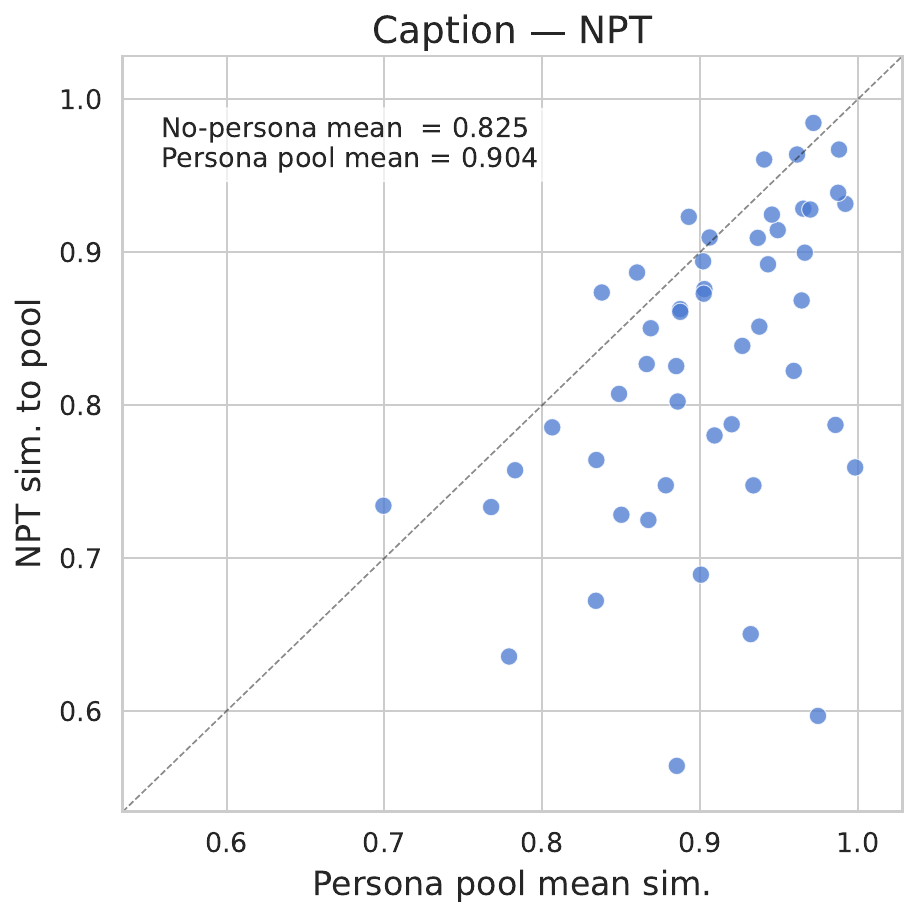}%
    \label{fig:no_persona_cap_npt}}
    \hfill
    \subfloat[Captions, NPNoT]{%
    \includegraphics[width=0.235\textwidth]{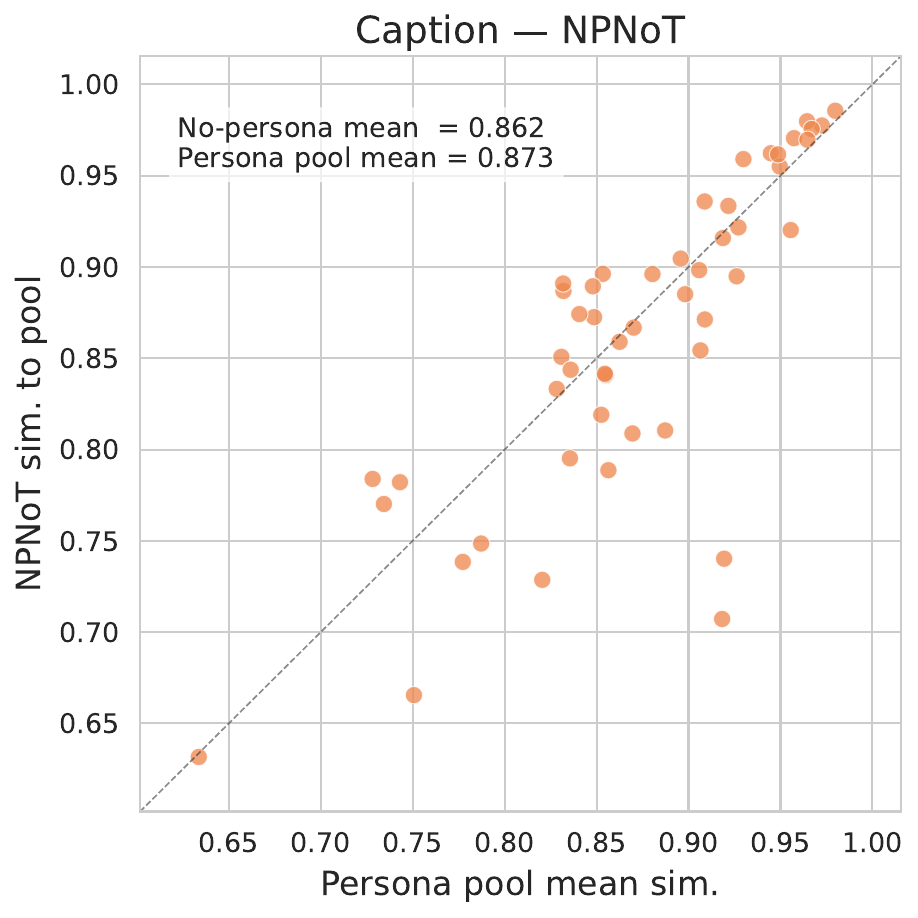}%
    \hspace{2pt}%
    \includegraphics[width=0.235\textwidth]{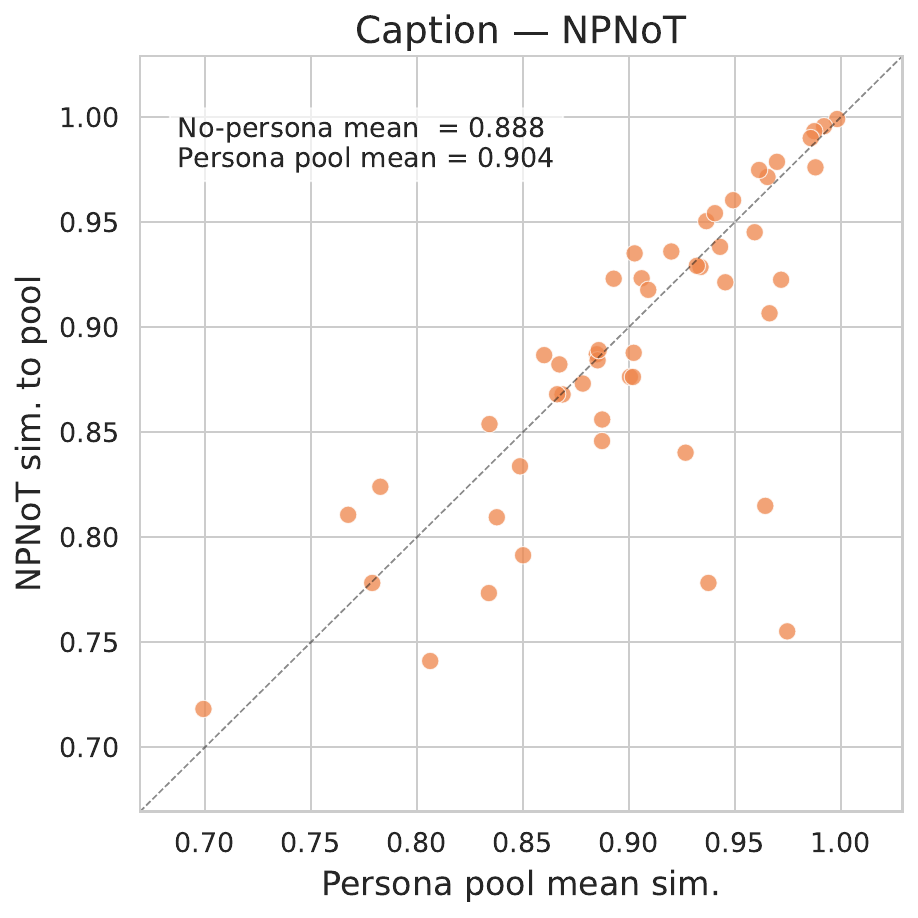}%
    \label{fig:no_persona_cap_npnot}}\\[6pt]
    \subfloat[Justifications, NPT]{%
    \includegraphics[width=0.235\textwidth]{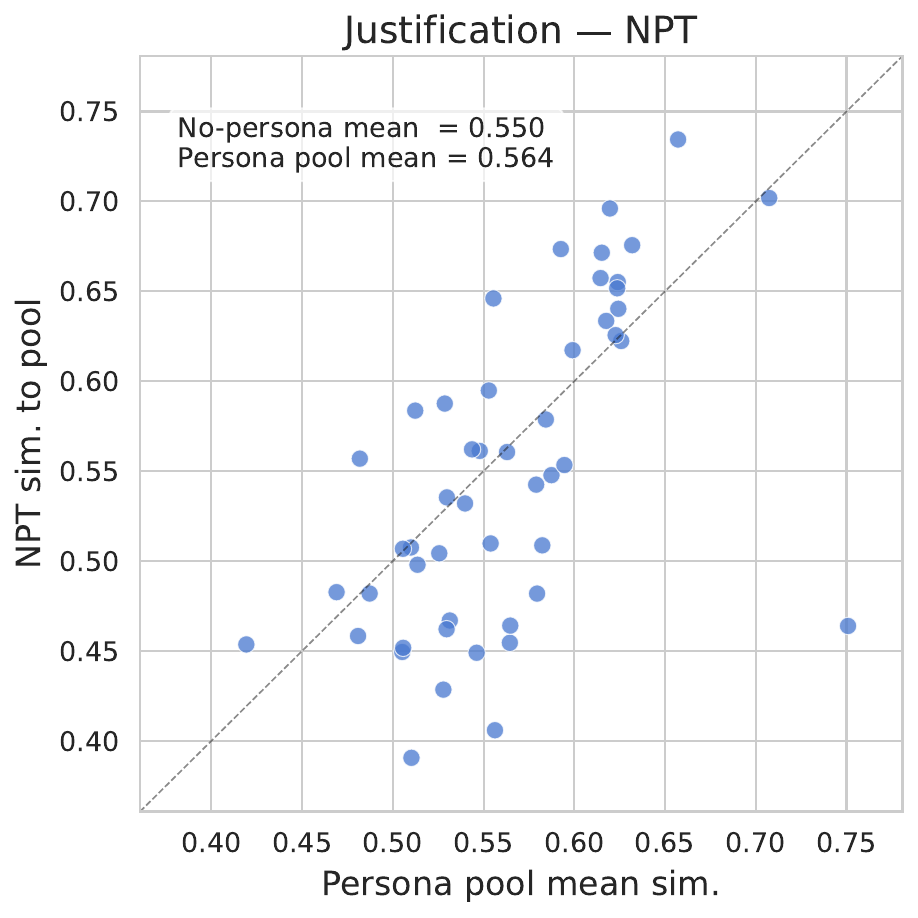}%
    \hspace{2pt}%
    \includegraphics[width=0.235\textwidth]{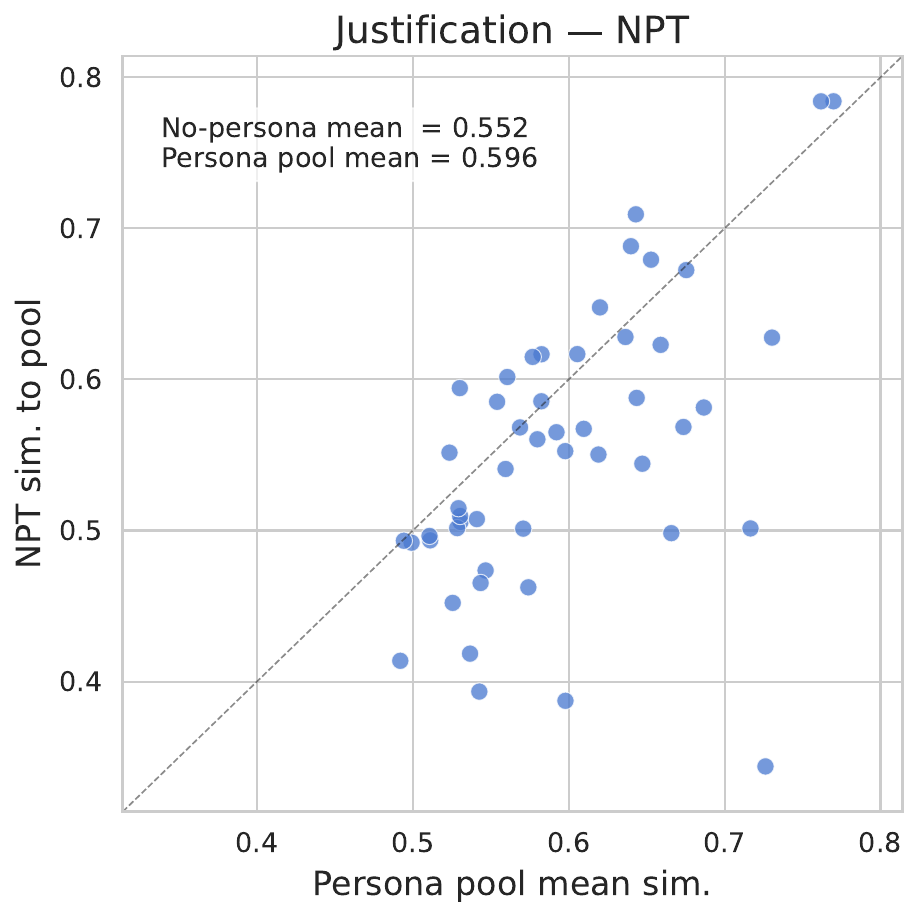}%
    \label{fig:no_persona_just_npt}}
    \hfill
    \subfloat[Justifications, NPNoT]{%
    \includegraphics[width=0.235\textwidth]{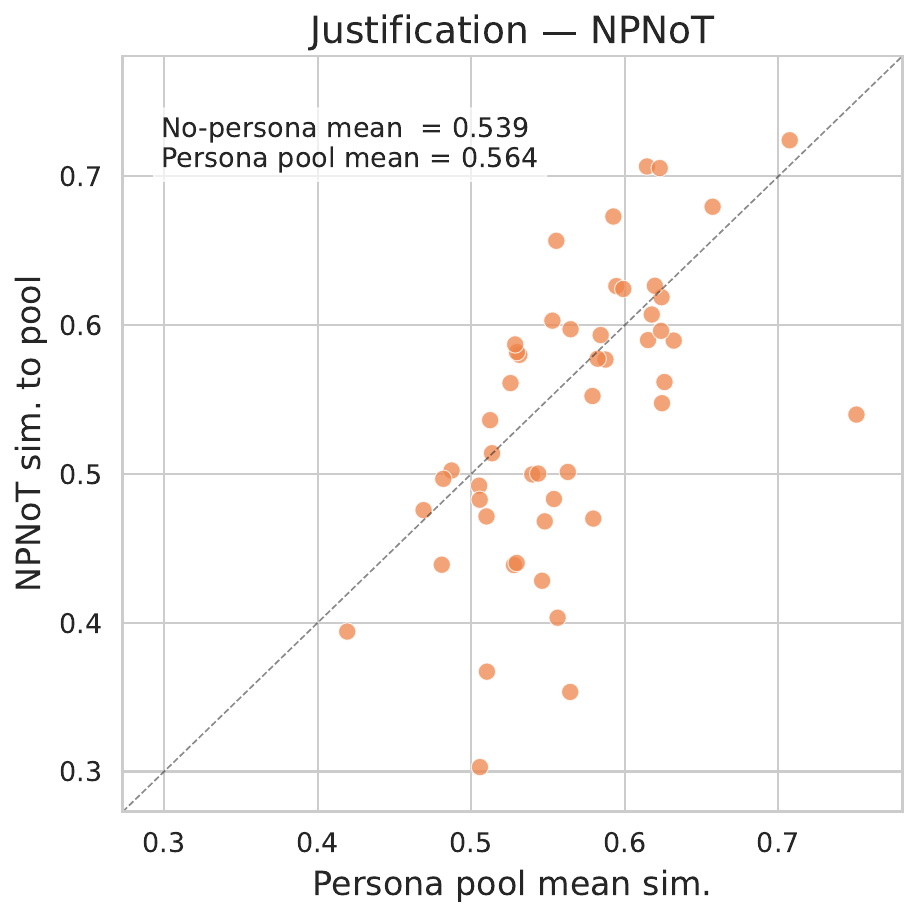}%
    \hspace{2pt}%
    \includegraphics[width=0.235\textwidth]{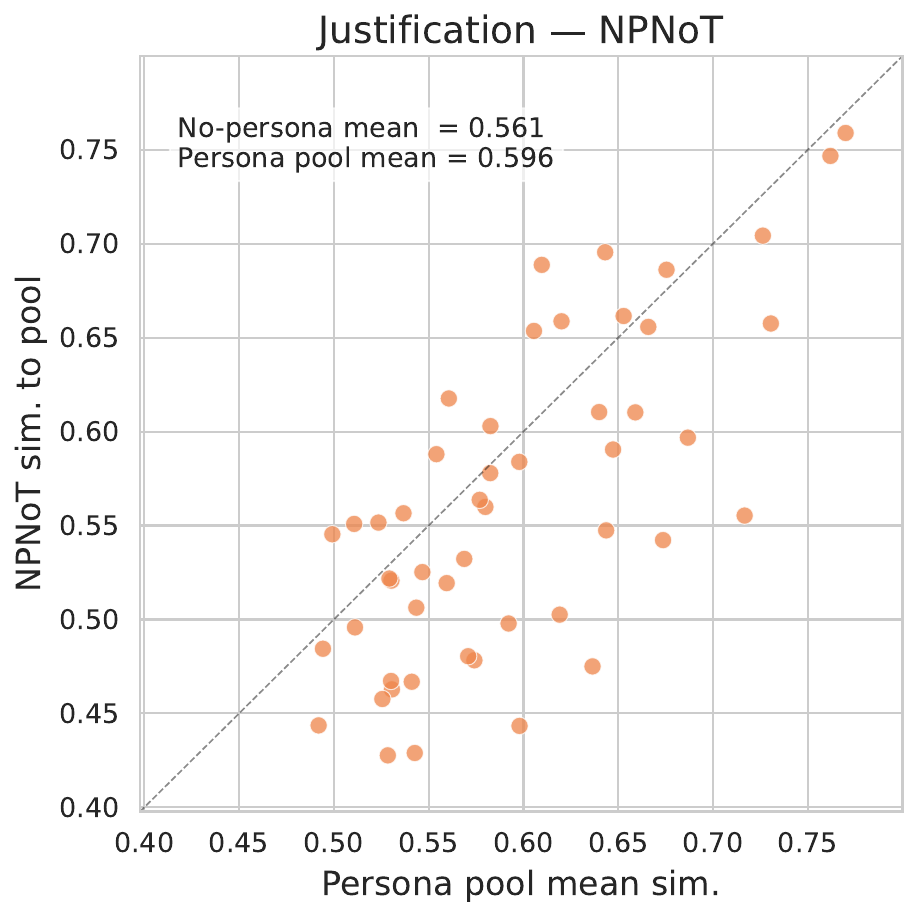}%
    \label{fig:no_persona_just_npnot}}\vspace{-0.4cm}
    \caption{Per-image cosine similarity of no-persona outputs to the persona-conditioned pool: (a) captions with extended reasoning (NPT), (b) captions without extended reasoning (NPNoT), (c) justifications with extended reasoning (NPT), and (d) justifications without extended reasoning (NPNoT). Within each panel, Qwen3-VL is shown on the left and Gemma4 on the right. Each point is one image; the dashed line marks $y=x$.}
    \label{fig:no_persona_sim}
\end{figure*}

\section{Persona \emph{vs.} No-Persona Effects}
\label{app:nopersona}

To contextualize the role of persona conditioning, we compare two no-persona variants (NPT and NPNoT), for each model, against the model's persona-conditioned pool using the same embedding-based cosine similarity described in Section~\ref{secWithPerStability}, computed separately for each image.

For each image $i \in I$, we consider two quantities: (i) the average cosine similarity among persona-conditioned outputs (persona-persona), and the cosine similarity between the no-persona output and the mean representation of the persona-conditioned outputs. Figure~\ref{fig:no_persona_sim} plots these for both models. Each point indicates how closely a no-persona response aligns with the persona pool relative to the internal agreement among persona-conditioned agents, allowing us to assess whether persona conditioning produces systematic shifts in generated language beyond the model's internal persona-conditioned variability.

Note that this comparison is structurally asymmetric: the persona-conditioned pool comprises $\approx 1{,}200$ annotations per image, whereas each no-persona setting contributes only a handful. The no-persona values should therefore be interpreted as indicative rather than statistically equivalent baselines.

For captions, both no-persona variants yield similarity levels close to the persona within-pool average (Qwen3-VL ---NPT: $0.865$; NPNoT: $0.862$; persona within-pool: $0.873$; Gemma4 --- NPT: $0.825$, NPNoT: $0.888$, within-pool: $0.904$). This is consistent with the strong convergence in descriptive grounding observed in Section~\ref{sec:results-wihin}, indicating that descriptions remain stable even without persona conditioning.

In contrast, justification similarities are lower overall (Qwen3-VL --- NPT: $0.550$; NPNoT: $0.539$; within-pool: $0.564$; Gemma4 --- NPT: $0.552$, NPNoT: $0.561$, within-pool: $0.596$), reflecting the greater semantic variability of interpretive framing relative to captions. In both models, the no-persona outputs remain broadly aligned with, yet slightly below, the persona within-pool average, consistent with the broader pattern observed in Section~\ref{sec:concordance}, where persona conditioning introduces structured variation primarily in justificatory language.

Beyond these mean differences, Figure~\ref{fig:no_persona_sim} shows that this pattern holds consistently across images for both models: caption similarities cluster tightly near the persona within-pool level, whereas justification similarities are both lower and more variable, reinforcing that persona effects primarily operate in interpretive language. 

These comparisons are consistent with the main finding that descriptive grounding is more stable than interpretive framing. Because the persona and no-persona conditions contain substantially different numbers of generations, however, they should be interpreted as contextual evidence rather than as a definitive test separating persona-conditioned effects from stochastic variation.

%% file: Appendixes/Append_TopicAnalysis.tex
\section{Topics Analysis}
\label{app:TopicsAnalysis}

\begin{figure}[t]
\centering
\subfloat[Qwen3-VL]{%
    \includegraphics[width=0.47\textwidth]
    {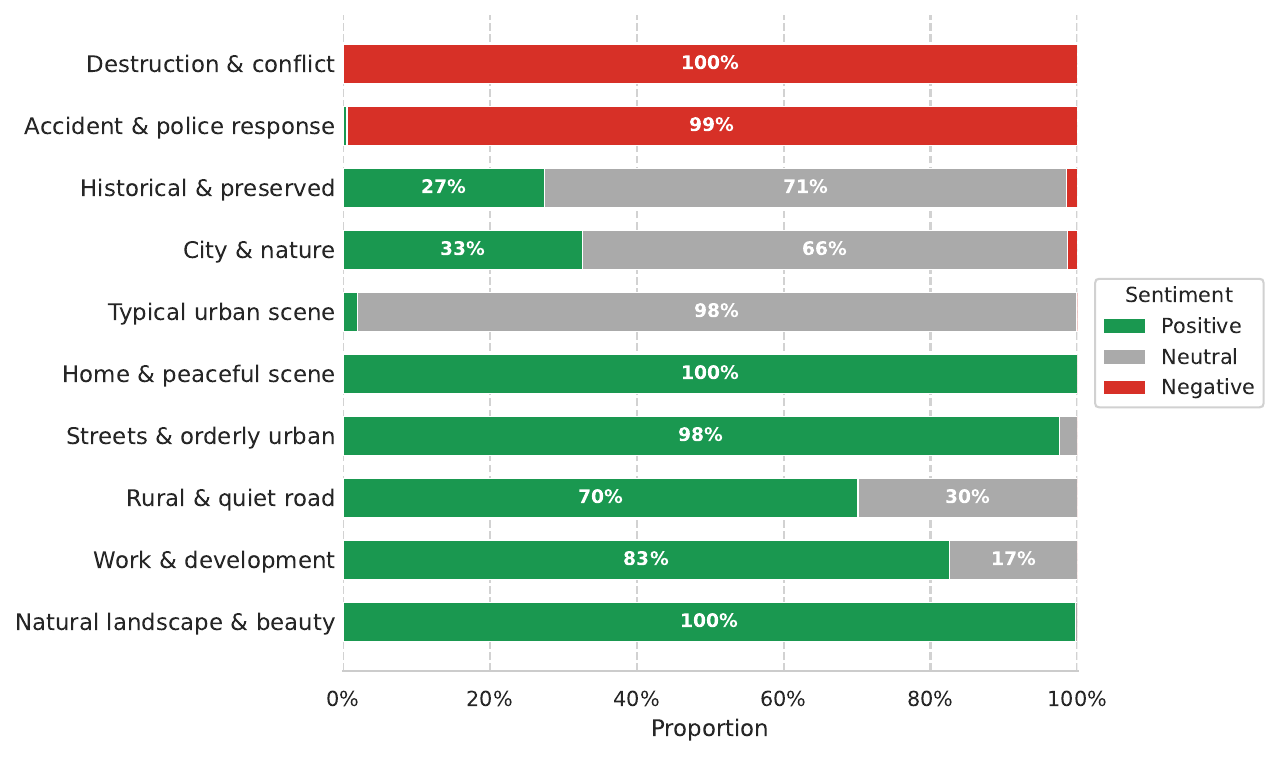}}
\hfill
\subfloat[Gemma4]{%
    \includegraphics[width=0.47\textwidth]
    {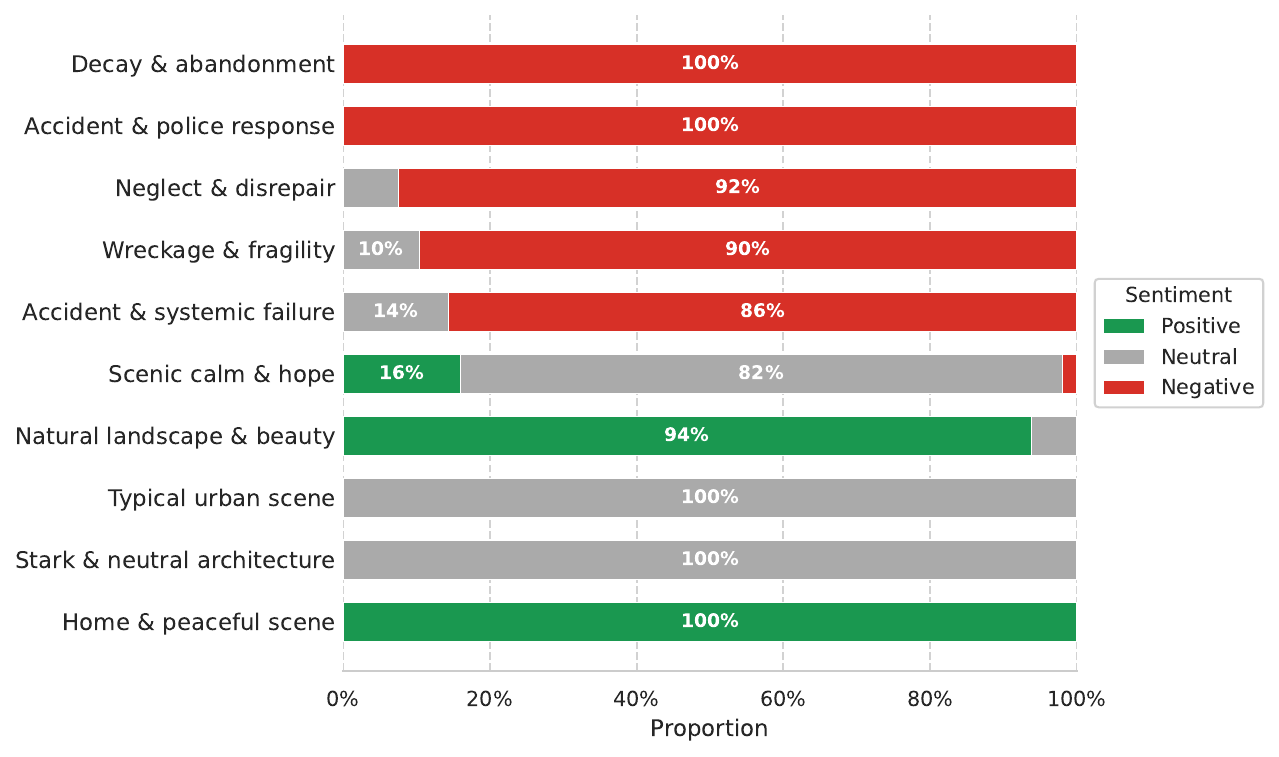}}
\vspace{-0.2cm}
\caption{Sentiment distribution across the top 10 justification topics
generated by persona-conditioned agents for Qwen3-VL and Gemma4.}
\label{fig:topic_sentiment}
\end{figure}

Figure~\ref{fig:topic_sentiment} shows that, in both models, justification topics align with sentiment in interpretable ways: accident, conflict, and decay topics skew negative; nature/beauty and warmth topics skew positive; and infrastructure or neutral/stark topics sit in between, reflecting ambiguity in how agents interpret ordinary urban development scenes.

%% file: Appendixes/Append_H_Prompts.tex
\begin{figure*}[!h]
\begingroup\color{gray}
\tiny
\noindent\textbf{[System prompt, persona condition]}
\begin{verbatim}
You are a person with the following demographic profile:
- Gender: {gender}
- Economic status: {economic_status}
- Political leaning: {political_spectrum}
- Personality archetype: {personality}

CRITICAL RULES:
1. You are this person. You are NOT an AI.
2. Evaluate the image exactly as this person would, based on your background.
3. Do not break character. Do not apologize for your views.

TASK — IMAGE SENTIMENT ANNOTATION:
Look at the image and respond with a JSON object and nothing else.

The JSON must have exactly these four keys:
  "sentiment"    : one of ['Positive', 'SlightlyPositive', 'Neutral', 'SlightlyNegative', 'Negative']
  "perceptions"  : a JSON array of 1-5 labels chosen EXCLUSIVELY from the
                   PERCEPTION VOCABULARY listed below — do NOT invent new labels
  "caption"      : a 1-2 sentence objective description of what is shown in the
                   image, written independently of your persona
  "justification": a single sentence in YOUR voice explaining your sentiment score

PERCEPTION VOCABULARY (593 labels — choose 1-5 that best match the image):
["\"Closed\" environment","\"Home\" feeling, something reminiscent of childhood",
 "\"Water mirror\"","Abandoned", ... ]

Example response:
{"sentiment": "Negative", "perceptions": ["Violence", "Debris/Destruction"],
 "caption": "A damaged street with rubble and broken glass scattered across the pavement.",
 "justification": "Looks like the same mess they always leave after those marches downtown."}

Return ONLY the JSON object. No markdown, no code fences, no additional text.
\end{verbatim}

\noindent\textbf{[System prompt, no-persona condition: the persona block and CRITICAL RULES are replaced by]}
\begin{verbatim}
You are an impartial image annotator. Evaluate each image objectively,
without any personal or demographic lens.
\end{verbatim}

\noindent\textbf{[User turn, both conditions: this text, followed by the image]}
\begin{verbatim}
What is the sentiment of this image?
\end{verbatim}
\endgroup
\vspace{-0.3cm}
\caption{Prompt templates. Placeholders in braces are filled with the persona's attribute levels. The perception vocabulary is truncated here (593 labels in full); line breaks in the vocabulary, example response, and no-persona prompt were added for layout.}
\label{fig:prompts}
\end{figure*}

\section{Prompt Templates}
\label{app:prompts}

Figure~\ref{fig:prompts} shows the prompt templates used for all annotations in both models. In the persona condition, the profile is specified through a direct second-person role instruction in the system prompt, with each placeholder filled with the corresponding attribute level (e.g., \texttt{Low income} or \texttt{High income} for economic status). The same annotation task and output instructions are used in the persona and no-persona conditions; only the persona-specific block and associated rules are replaced in the no-persona condition. The prompt explicitly requests a caption written independently of the persona and a justification in the persona's own voice, a design choice discussed in the Limitations section.